\documentclass[]{interact}

\usepackage{epstopdf}% To incorporate .eps illustrations using PDFLaTeX, etc.
\usepackage[caption=false]{subfig}% Support for small, `sub' figures and tables
\usepackage{amsmath,amssymb,amsfonts}
\usepackage{caption}
\usepackage{subfig}
\usepackage{graphicx}
\usepackage{textcomp}
\usepackage{xcolor}
\usepackage{array}
\usepackage{apacite}
\usepackage{makecell}
\usepackage{xurl}
\usepackage{cleveref}
\usepackage[longnamesfirst,sort]{natbib}% Citation support using natbib.sty
\bibpunct[, ]{(}{)}{;}{a}{,}{,}% Citation support using natbib.sty
\renewcommand\bibfont{\fontsize{10}{12}\selectfont}% To set the list of references in 10 point font using natbib.sty

\theoremstyle{plain}% Theorem-like structures provided by amsthm.sty

\theoremstyle{definition}

\theoremstyle{remark}

\begin{document}
\pagestyle{plain} % You can also use "headings" or "empty" depending on your preference

\pagenumbering{arabic} % You can use "roman" or "Roman" for Roman numerals

%\articletype{ARTICLE TEMPLATE}% Specify the article type or omit as appropriate

\title{GNN-based Passenger Request Prediction}

\author{
\name{Aqsa Ashraf Makhdomi \thanks{ CONTACT Aqsa Ashraf Makhdomi. Email: makhdoomiaqsa@gmail.com} and Iqra Altaf Gillani }
\affil{Department of Information Technology, NIT Srinagar }
}

\maketitle

\begin{abstract}
Passenger request  prediction  is  essential  for  operations planning,  control,  and  management  in ride-hailing platforms. While the demand prediction problem has been studied extensively, the Origin-Destination (OD) flow prediction of passengers has received less attention from the research community. 
This paper develops a  Graph Neural Network (GNN) framework along with the Attention Mechanism to predict the  OD flow of passengers. The proposed framework exploits various linear and non-linear dependencies that arise among requests originating from different locations and captures the repetition pattern and the contextual data of that place. Moreover, the optimal size of the grid cell that covers the road network and preserves the complexity and accuracy of the model is determined.
Extensive simulations are conducted to examine  the  characteristics of our proposed approach and its various components. The results show  the superior performance of our proposed model compared to the existing baselines.
\end{abstract}

\begin{keywords}
Ride-hailing; Route recommendation; Demand prediction; OD prediction; GNN; Context-aware data
\end{keywords}

\section{Introduction}

\label{sec:Introduction}

The rapid growth of GPS-enabled services and location-based sensors has resulted in an enormous volume of geo-tagged data which provides information about passenger mobility patterns and vehicular movement. This data about passenger arrival and departure from different locations can be analyzed and the patterns can be exploited among them in order to predict the future areas that will fetch more requests.
This will help the ride-hailing platforms in assigning the requests of nearby passengers to the drivers. It will ultimately decrease the waiting time of passengers and the cruising distance of drivers without having a passenger in the vehicle. The decrease in cruising distance of vehicles results in more profit for the platform as the passengers are fetched earlier on the route. 
It further has a positive impact on the environment as there is a reduction in the emission of greenhouse gases, %considering 
owing to the decrease in distance travelled by the vehicle. %as the quantity of greenhouse emissions is reduced.
Thus the prediction of requests  enhances the service quality of 
ride-hailing platforms and contributes to a greener environment. %reduces the emission of greenhouse gases, which contributes to eco-friendly rides.  
This is an important area of research and the output of these prediction techniques is used as input by various route recommendation \cite{Garg:ACMKDD_2018}  %\textcolor{blue}{[[IAG: Give a citation here as well]]} %\cite{Garg:ACMKDD_2018} 
and matching algorithms \cite{Gao:IEEETrans_2022}. 
It has gathered significant attention from researchers and a number of works like \cite{Wang:IEEEConf_2021,Jin:ElsevierTransp_2020} have been done that have predicted the areas with more passenger demand. 

Most of the works \cite{Wang:IEEEConf_2021,Jin:ElsevierTransp_2020} that have been done in this direction have predicted the areas that will fetch more requests, which is also called demand prediction.  However, the mobility patterns of passengers can be better analyzed if the origin (the place from where the request came), as well as the destination (the place to which the request is headed) of requests, can be predicted simultaneously and not only their origin.  This is particularly useful for ridesharing platforms (platforms that enable multiple passengers to share a single ride)  where the efficient pairing of passengers can be done if the origin and destination   of passengers are known beforehand. It is a new area of interest and various models have been proposed to predict the Origin-Destination (OD) pair. Some of the studies have 
explored the application of statistical models \cite{Cho:KDD_2011} for prediction, but their accuracy is below $50\%$ which can be attributed to the complex spatio-temporal dependencies inherent in request data. %These works have mainly used deep-learning models to exploit the spatial and temporal dependencies among requests and predicted their origins and destinations accordingly. 
These dependencies have been effectively captured by employing a range of deep-learning models \cite{Schaller:ElsevierTransport_2021,Seo:NIPS_2018,Wang:ACMKDD_2019,Wang:ACMTrans_2022,Shen_IEEEAccess:2022}. 
Among them, Graph Neural Networks (GNNs) \cite{Seo:NIPS_2018,Wang:ACMKDD_2019,Wang:ACMTrans_2022,Shen_IEEEAccess:2022}  have gained prominence over Convolutional Neural Networks (CNNs) due to their capability to account for non-euclidean distances among neighboring data points. %In this direction, various works have used GNN and its variants \cite{Seo:NIPS_2018,Wang:ACMKDD_2019,Wang:ACMTrans_2022,Shen_IEEEAccess:2022} for OD prediction. 
All of these models capture the linear trends in data that occur due to the repeating schedules of passengers. However, apart from linear trends, there can be non-linear trends in data.

Our proposed model, in addition to these linear patterns, also aims to capture non-linear patterns that may result in dependencies between requests.
It uses GNN to model the underlying road network with different linear and non-linear dependencies among the ride-hailing requests. %These dependencies can arise due to linear trends in data that have been captured by the previous models \cite{Seo:NIPS_2018,Wang:ACMKDD_2019,Wang:ACMTrans_2022,Shen_IEEEAccess:2022} and the non-linear trends which appear mainly due to the contextual data and the travelling behavior of people. 
%Like the previous studies \cite{Wang:ACMTrans_2022,Shen_IEEEAccess:2022}, t
The linear trends in data are determined through the data from the same or consecutive hours of the preceding time frames.  
Whereas the non-linear patterns in data are determined by taking the aggregate information from the data of the preceding hours. This data provides insights into the behavioral patterns of people and describes the contextual information of that place. For instance, if the requests from the preceding hours are low in quantity,  it reveals that either the area is sparse in requests or some event has occurred that has reduced the flow of requests to that area. Moreover, these dependencies also reflect the travelling behavior of people and thereby indicate the re-appearance of requests after a period of time. For instance, requests start to recur after office hours of people are finished or they have completed the time spent outside their homes. These dependencies are captured by analyzing the data from previous hours and determining the time frame after which requests tend to re-appear. %  Thus the data from previous hours provides a mechanism to deal with non-linear dependencies that describe the patterns apart from the regular repeating trends that are found in the existing studies.}
\vspace{1mm}

%\textcolor{bluWhile demand prediction has garnered significant attention from researchers, there remains a relatively limited body of work focused on the Origin-Destination (OD) prediction of ride requests.These works have mainly used deep-learning models to exploit the spatial and temporal dependencies among requests and predicted their origins and destinations. Some of the studies have  explored the application of statistical models \cite{Cho:KDD_2011} for prediction, but their accuracy is below $50\%$ which can be attributed to the complex spatio-temporal dependencies inherent in request data. These dependencies have been effectively captured by employing a range of deep-learning models. Notably, GNNs have gained prominence over Convolutional Neural Networks (CNNs) due to their capability to account for non-Euclidean distances among neighboring data points.  In this direction, various works have used GNN and its variants \cite{Seo:NIPS_2018,Wang:ACMKDD_2019,Wang:ACMTrans_2022,Shen_IEEEAccess:2022} for OD prediction. All of these models capture the periodic trends in data accurately. However, apart from periodic trends, there can be non-periodic trends in data.} 

Grid size is a crucial factor to consider when modelling these dependencies since it determines the number of spatial and temporal neighbors at a given time. When the grid size is large, the neighborhood count decreases which results in the inability to account for spatio-temporal dependencies. On the other hand, small grid sizes make it necessary to retain microscopic features, which ultimately leads to an increase in the complexity of the model. Thus, the appropriate grid size which captures the passenger mobility accurately in a time-bound manner is a parameter that needs to be determined. 

    Our major contributions can be summarized as follows:
    \begin{itemize}
        \item 	We model the ride-hailing network as a graph and use GNN to capture the complex spatio-temporal dependencies among the requests that arrive from different locations and varying time frames. 
\item We determine the temporal functioning of ride-hailing platforms and explore non-linear patterns that appear within the request sequences. 
\item %The proposed model 
The proposed model displays that apart from the periodic trends in data, the dependencies among requests can arise due to the movement and behavioral patterns of people.
%We capture the movement patterns of people due to their travelling behavior and determine the contextual data of the place through the use of a non-linear layer. }% which enhances the predictive power of the model.}%

%\item \textcolor{blue}{The proposed model captures the movement patterns of people and tells about their travelling behavior which provides a more comprehensive understanding of how demand evolves over time.}
%The temporal aspects of data also reveal the contextual information of the place and provides ride-hailing platforms with an indication of locality behavior of the place over a particular time frame. Incorporating contextual data about the specific time of day or week enhances the predictive power of our model. This analysis enables us to capture movement patterns of people and tells about their travelling behavior which provides a more comprehensive understanding of how demand evolves over time.}
\item	We determine the optimal size of the grid cell considering the complexity and accuracy of the model.

%We analyze the temporal non-linearities in the ride-hailing request sequences and capture the contextual data of the place.
\item	 We conduct extensive simulations using a real-world dataset to empirically validate the effectiveness of our proposed model. The superior performance demonstrated in these simulations showcases the practical utility and relevance of our approach in ride-hailing platforms. %Our model's ability to outperform existing methods highlights its potential to enhance user experiences and optimize operational efficiency in the ride-hailing industry.%Extensive simulations conducted on the real-world dataset demonstrate superior performance by our proposed model.
    \end{itemize}

The rest of the paper is organized as follows: In Section \ref{sec:related work}, a review  of the existing work done related to the request prediction is presented. In Section \ref{sec:preliminaries}, the preliminaries required to understand our proposed model are described. In Section \ref{sec:systemmodel} the details of the proposed model are discussed. In Section \ref{sec:implementation}, the dataset, evaluation metrics, results, and comparison of algorithms is discussed. Section \ref{sec:apps} highlights the key applications of the proposed model in ride-hailing industry. Finally,  Section \ref{sec:conclusion} concludes our work and highlights its key contributions.

\section{Related Work}
\label{sec:related work}
Ride-hailing services like Ola and Uber generate an enormous volume of data, which includes information about trajectories, geo-tagged check-ins, and ride sources and destinations. The patterns among data can be learned and understood for the advanced development of these services in recommending optimal routes \cite{Garg:ACMKDD_2018} or providing matching algorithms \cite{Gao:IEEETrans_2022} to drivers and passengers. Various machine learning and deep learning-based models have been used to understand the patterns among data and utilize those patterns for future prediction of requests \cite{Wang:ACMTrans_2022,Han_2004}. 
These prediction algorithms for ride-hailing platforms work in two-fold directions: predicting the demand at nodes and predicting the OD pair of requests. Demand prediction methods predict the number of requests that will arrive at a node and the OD-based prediction methods predict the number of requests that will arrive between a specified origin and destination pair of vertices. 

%These works have evolved from pure time series based models \cite{} to the models exploiting spatial and temporal dependencies \cite{}. The time series based models exploit the temporal patterns in data and based on these predict the future  areas that will have more requests. Since ride-hailing requests depend upon both the spatial and temporal dependencies, some of the recent research works \cite{} have started to exploit both these dependencies by using various machine learning and deep learning based approaches. 

Various works have been in the direction of demand prediction. 
These works have evolved from pure time series-based models \cite{Han_2004,Lippi:IEEETrans-2013} to the models exploiting spatial and temporal dependencies \cite{Wang:IEEEConf_2021,Jin:ElsevierTransp_2020}. %,%Feng:IEEETrans_2021}. 
The time series-based models exploit the temporal patterns in data and based on these predict the future  areas that will have more requests. Since ride-hailing requests depend upon both the spatial and temporal dependencies, some of the recent research works have started to exploit both these dependencies by using various machine learning and deep learning-based approaches \cite{Wang:ACMTrans_2022,Han_2004}. %\cite{Wang:IEEEConf_2021,Jin:ElsevierTransp_2020}. %,Feng:IEEETrans_2021}. %These works have developed from pure time-series based models \cite{Han_2004,Lippi:IEEETrans-2013}
%  \cite{Yuan:ACM_2011},  \cite{Li:IEEECon_2011}, \cite{Miao:IEEETrans_2019} \cite{Gammelli:Elsevier_2020}. 
%to the models exploiting spatio-temporal dependencies \cite{Wang:IEEEConf_2021,Jin:ElsevierTransp_2020,Wang:ACMKDD_2019,Feng:IEEETrans_2021}. %\cite{Geng:aaai_2019}\cite{Zhang1:IEEETrans_2021} The spatio-temporal based models exploit the dependencies that can arise due to repetition pattern among requests or the place from where they are originating.  These works are mostly based on Graph neural networks and Recurrent neural networks.
 
Recently, OD prediction has arisen as a potential topic among researchers as it can enhance the functionality of ride-hailing platforms, in particular ridesharing systems. Various works have been done that have used GNN  %, RNN and CNN \cite{Lai:ACMSIGIR_2018} %lstnet 
based models to predict the origin and destination of requests \cite{Hamilton:neuips_2017,Wang:ACMTrans_2022,Wang:ACMKDD_2019,Shen_IEEEAccess:2022}. In this direction, Hamilton \textit{et al.} \cite{Hamilton:neuips_2017} proposed  a general induction framework, GraphSAGE, that used the attribute information to create node embeddings for the vertices of the graph. However, their proposed model only focused on spatial dependencies and did not capture the temporal patterns in the data. 
Wang \textit{et al.} \cite{Wang:ACMKDD_2019} proposed a grid embedding-based multi-task learning framework, where the grid embedding models the spatial dependencies that can arise among requests from different areas, and LSTM-based multi-task learning framework captures the temporal dependencies in the data.
 However, their proposed model did not consider the direction of the flow of requests and they considered the requests that originate at node $v$ or are destined to node $v$ as the same. This could affect the performance of the system considerably. For instance, %if the request in morning rush hours is from the residential area to the 
if the requests are moving from their homes to shopping malls the direction of movement of requests is an important pattern and its ignorance can affect the system's performance significantly.
In order to overcome the above problems, Wang 
 \textit{et al.} \cite{Wang:ACMTrans_2022} proposed a representation learning-based OD prediction model that leveraged the spatial and temporal dependencies through the use of three types of neighbors: forward, backward, and geographical. Their proposed model took the directed nature of requests into account and accordingly considered the neighbors as forward or backward based upon their precedence of request. %\textit{et al.} proposed a multi-graph attention network  \textit{et al.} proposed a federated learning based frameowrk 
 Apart from these models, other works \cite{Ke:ElsevierTranport_2021,Shen_IEEEAccess:2022,Hu:IEEETVT_2023} have been done that have used multi-graph attention and federated learning-based models to find the dependencies among requests %using the data from previous hours or days 
 and accordingly predicted the origin and destination of future requests.
 
 All of the above works model the dependencies that arise due to the repeating trends in data. However, apart from these recurring trends, there can be dependencies due to the non-recurring patterns in data.    %do not model the dependencies that appear due to non-repeating trends in data 
 Our proposed model tries to overcome the above limitations and capture the spatial and temporal dependencies, in particular the dependencies that appear due to non-repeating patterns like contextual data or the travelling behavior of people. It is based on the architecture proposed by \cite{Wang:ACMTrans_2022} and \cite{Shen_IEEEAccess:2022}. %We propose to capture these dependencies accurately in a timely manner 
 %Seo et al. [19] build a model called GCRN to generalize the classical RNN to structured data by an arbitrary graph, which can be used to predict sequences of structured data. However, this method is incapable of modeling the transferring relationship between areas either because they have not taken the semantic neighbors into consideration.

%This is a new research area and the work is going on in this direction. Our proposed model also predicts the origin and destination of requests by using GNN and is based on the architecture of \cite{Wang:ACMTrans_2022}.
%Wang \textit{et al.} proposed a GNN based passenger mobility prediction model which analyzed the relationship among different requests through GNN. Our proposed model is also based on their architecture. We propose to predict the demand as well as origin, destination pair of ride-hailing requests by using GNN.  

\section{Preliminaries}
\label{sec:preliminaries}
In this section, we will discuss the preliminaries associated with our proposed model. 

\textit{Grid}

\begin{figure}[h]
    \centering
    \includegraphics[width=0.35\textwidth]{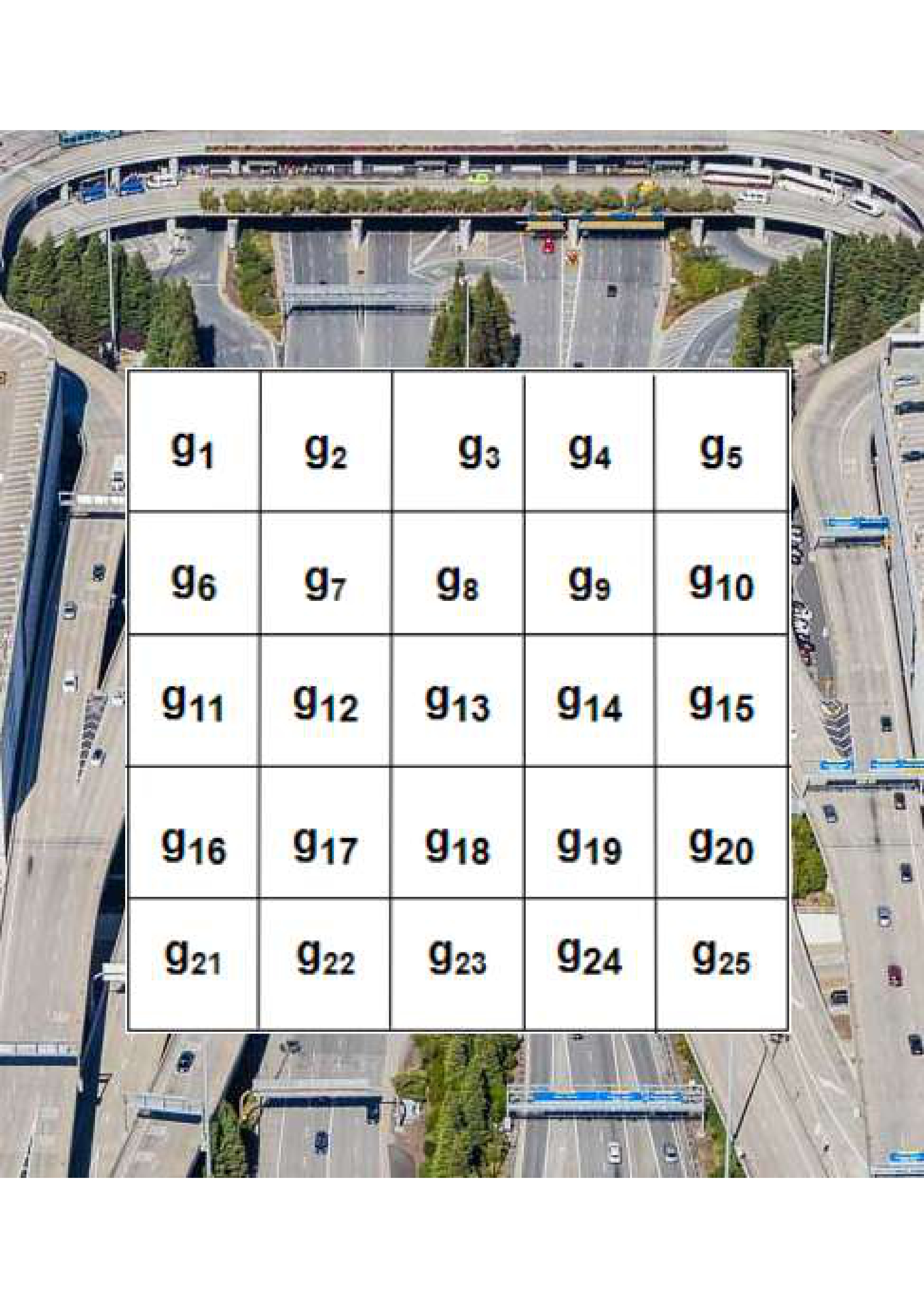}
    \caption{Road network represented in the form of a grid}
    \label{fig:mesh1}
\end{figure}

\begin{figure}[h]
\vspace*{-10mm}
    \centering
    \includegraphics[width=0.5\textwidth]{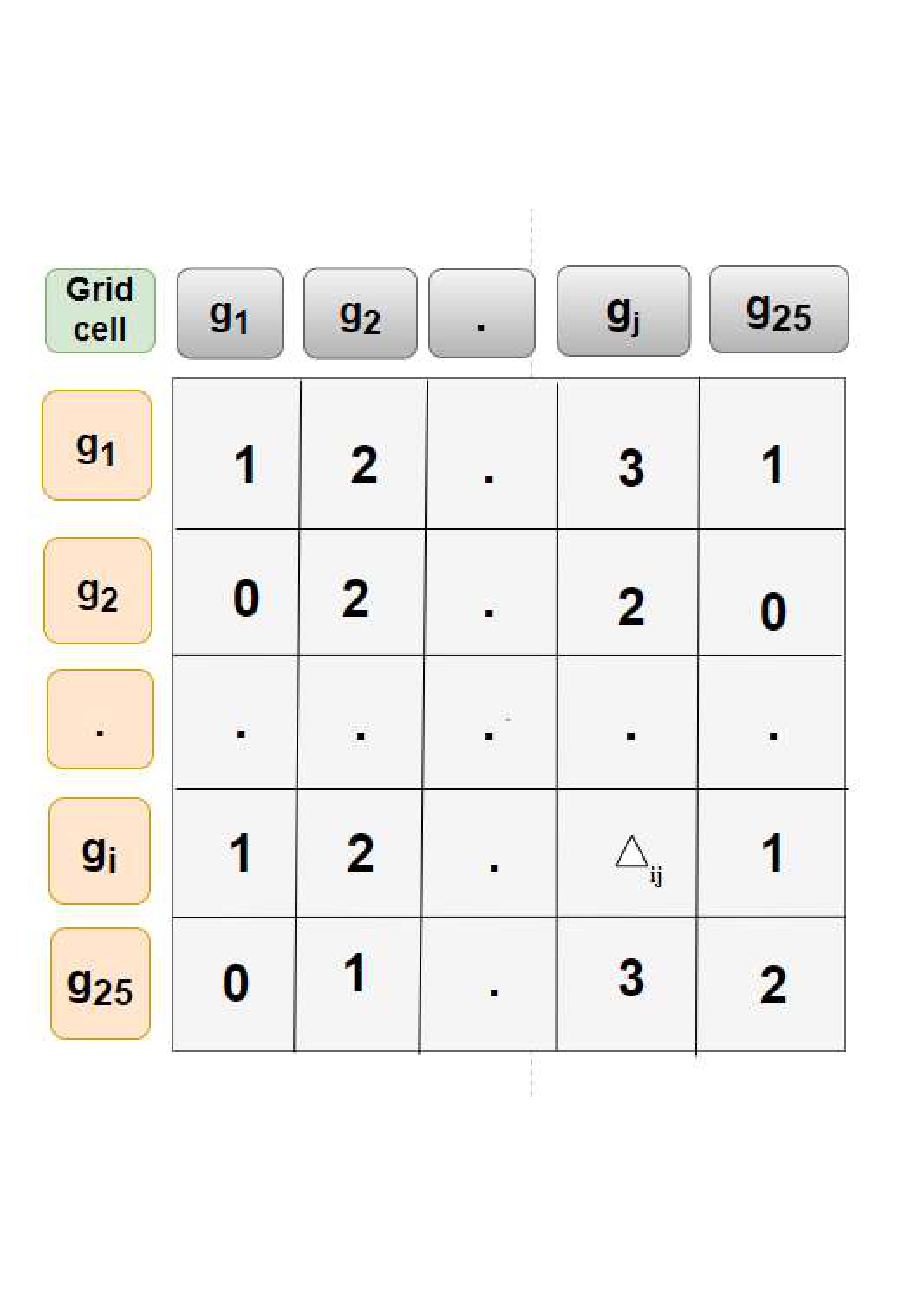}
    \vspace*{-8mm}
    \caption{OD Matrix of the grid}
    \label{fig:mesh2}
\end{figure}

Our proposed model assumes that the entire city is divided into $n$ non-overlapping grid cells, denoted by $g=\{g_1,g_2,...,g_n\}$. %Each grid cell is defined by its starting coordinate and its ending coordinate.
The grid cell represents a 
geographic area and is defined by its latitude and longitude coordinates. The distance between adjacent grid cells is measured between their central (middle) points.
For each cell of the grid, we predict the number of requests  that originate from there and are headed toward other cells.
Figure \ref{fig:mesh1} shows the example of a road network divided in terms of grid cells. The grid has $25$ cells and each cell $i$ of the grid is defined by its grid ID $g_i$ where $i \, \epsilon \, [1,25]$. 
Our proposed model predicts the number of requests that can arrive between any two cells of the grid.
%The number of requests that can arrive between two cells of grids are 
Their value is stored in the OD matrix, where the rows and columns of the matrix denote the grid cells and the entry of the matrix represents the number of requests that can arrive between those cells. Figure \ref{fig:mesh2} shows the OD matrix corresponding to the grid described in Figure \ref{fig:mesh1}. As can be seen through Figure \ref{fig:mesh2}, each entry of the matrix contains the number of requests ($\Delta_\mathrm{ij}$) that can arrive between the $i^\mathrm{th}$ cell of the grid ($g_i$) and the 
$j^\mathrm{th}$ cell of the grid ($g_j$), where $i,j\, \epsilon\,[1,25]$.

%The grid cell represents a geographic area and is represented by its latitude and longitude coordinates. The distance between adjacent grid cells is measured between their central points. 

%part of road 

\textit{Time Slots}

We split time into $24$ distinct slots $t=\{t_1,t_2,...,t_\mathrm{24}\}$ where $t_i$ represents the time between the $i^\mathrm{th}$ and $(i-1)^\mathrm{th}$ hour. For instance, $t_1$ corresponds to the time between $12:00$ A.M. and $1:00$ A.M..

%the difference between consecutive slots is $1$ hour. %constant. %In our proposed architecture, time is divided into $24$ slots and the difference between consecutive slots is $1$ hour.

\textit{Spatio-temporal dependencies}

The requests in ride-hailing platforms have temporal dependencies as the studies have found that there are regular time-based patterns among requests that could be exploited for further prediction. For instance, during morning hours, the requests follow the pattern of having a destination at the office, and during the evening hours, requests originate from the office. These patterns can be used to predict the future origin and destination of requests.

Requests also follow spatial dependencies and depend upon the inflow and outflow of requests from the following areas, i.e., if the neighborhood region has high requests, it is highly probable that the current region will have more requests. Similarly, if a region is sparse in requests, its subsequent regions will certainly have few requests.

\section{System model}
\label{sec:systemmodel}
\begin{figure}[h]
\vspace*{-35mm}
\centering
    \includegraphics[width=0.8\textwidth]{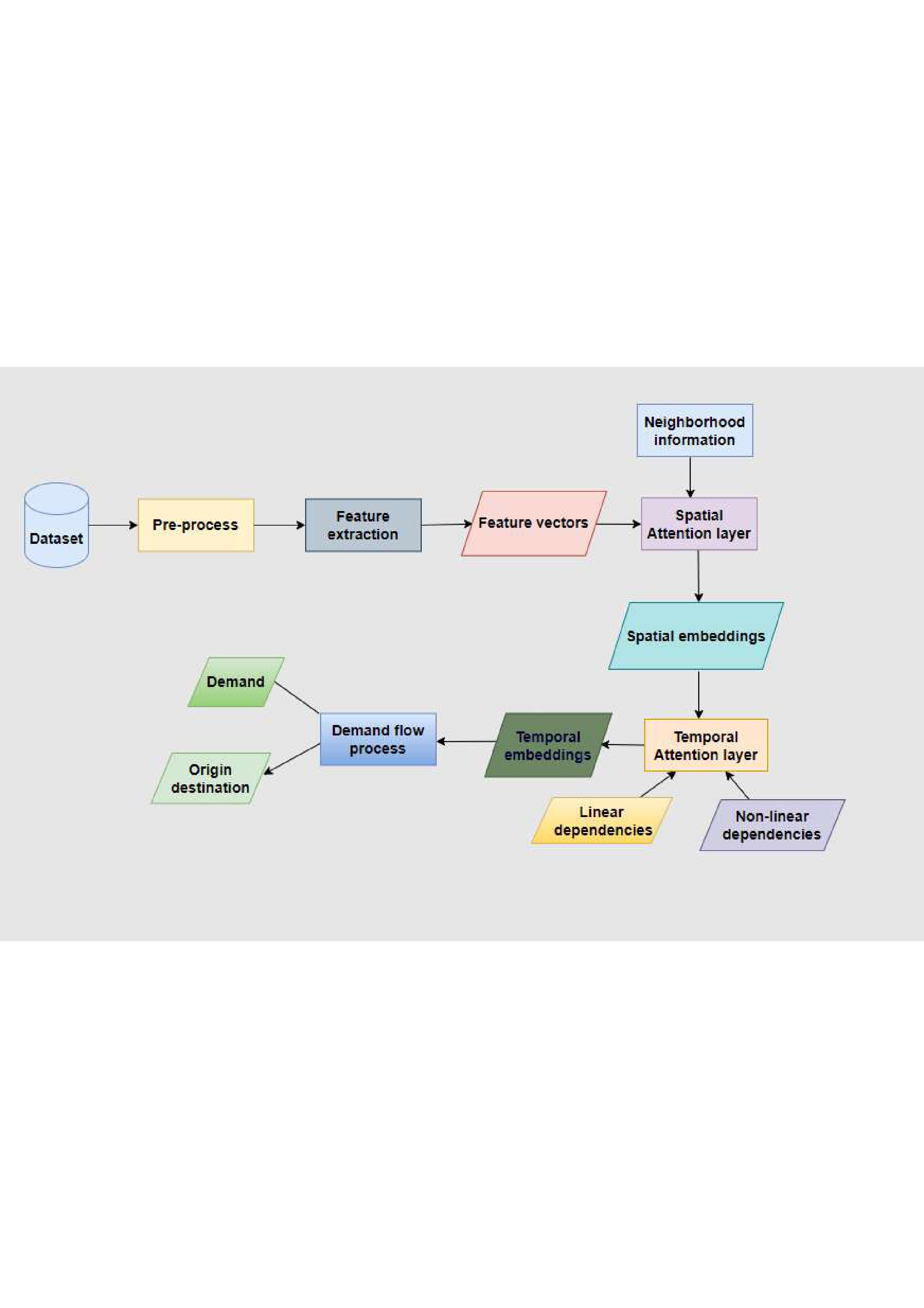}
    \vspace*{-30mm}
    \caption{Framework of our proposed model}
    \label{fig:framework}
\end{figure}
%Vector that represents the embedding of node $v_i$ concatenated with its neighbor $v_j$ that  has been passed through a pre-weighted aggregator
In this section, we detail the working of our proposed architecture.  
As can be seen through Figure \ref{fig:framework}, input to our proposed model is a dataset that contains the request sequence $R$$=$$\{r_1,r_2,...,r_m\}$ of ride-hailing platforms, where $r_i$ represents a particular request and $m$ denotes the total number of requests. %that arrives in ride-hailing platform. 
Each request is represented as $r_i$$=$$<$$r_{s_i},r_{d_i},r_i^\mathrm{t'}$$>$, where  $r_{s_i}$ and $r_{d_i}$ determine the  source and destination of request $i$ respectively, %They are correspondingly represented  in the form of latitude/longitude pairs. 
and $r_i^\mathrm{t'}$ represents the time at which the request $i$ is made. This data is pre-processed and divided into $24$  slots of $1$-hour for each day and a graph $G$$=$$(V,E,\Delta)$ is created for each time slot. The vertices $V$ of the graph denote the  grid cells %which were represented
and the edges $E$ represent the interconnections between different grid cells.
 %The rows/columns of the OD matrix represent the nodes $V$ of the graph and each of them describes the cells of grid (see Figure \ref{fig:mesh1}), and the edges $E$ represent the interconnection between different grid cells.  
This graph is a complete graph as requests can arrive between any pair of vertices. The adjacency matrix of the graph $G$ is represented by the OD matrix shown in Figure \ref{fig:mesh2}. %As can be seen through Figure \ref{fig:mesh2},
Each edge of the graph $G$ is associated with a weight $\Delta_\mathrm{ij}$ that determines the number of requests between the grid cells $g_i$ and $g_j$. %  (see Figure \ref{fig:mesh2}). 
If there are no requests between the two vertices, the corresponding weight is set as $0$. %$\Delta$ is a weight learned for OD prediction tasks and it represents the number of passengers between any of the two vertices of the graph. The demand prediction methods predict the demand at a node % and our proposed model tries to learn it and predict it.  
%\textcolor{red}{This weight $\delta$ is learned and predicted by our proposed model.}
The output of the pre-processed stage is a sequence of graphs $G$$=$$\{G_1,G_2,...,G_\mathrm{24}\}$, where $G_i$ represents the requests that arrive in the $t_i^\mathrm{th}$ time slot i.e, all the requests that arrive between $(i-1)^\mathrm{th}$ and $i^\mathrm{th}$  hour of the day.  
%$G_i$ represents the request sequence $R_i=\{r_\mathrm{i1},r_\mathrm{i2},...,r_\mathrm{iz}\}$where  $ i\,\epsilon \,[1,24]$. $R_i$ corresponds to the set of requests that arrive in time slot $t_i$ i.e, all requests that arrive between $(i-1)^\mathrm{th}$ and $i^\mathrm{th}$  hour. %and it denotes  the time slot arrive within a particular time slot in set $T$. 
%where $r^t_i$ lies between $T$ 
%where the requests $r_i \, i\,\epsilon \,[1,z]$ arrive within a particular time slot in set $T$. 
%$$ T\leq r^t_i < T+1, \,i\, \epsilon \, [1,z]%,T\, \epsilon \, [0,23] \,
 %$$ $T$ dent
% i.e,  graph $G_i$ corresponds to the requests that arrive within the time period $t_\mathrm{i-1}$ and $t_i$.  %at the     %$$r=<r_s,r_d,r^t>$$ and $$ \forall i T<r_i^t<T+1, \;T\, \epsilon \, [0,23]$$ i.e, each graph corresponds to the requests that arrive within the time period $T$ and $T+1$.  %at the corresponding time slot. 
This graph sequence is generated for all the days in the dataset. %The graph $G$ represents the flow of requests and it is different for all the time slots. The other graph $D$ is used to represent the distance between different road intersections and it is constant across all the time slots.

\begin{table}
\vspace{-5mm}
\begin{center}
%\tiny
\scalebox{0.7}{
%\resizebox{\textwidth}{!}{     
\begin{tabular}{ |l|l| } 
 \hline
 Notation & Description  \\ \hline
 $g$ & Grid  \\ 
 \hline
 $g_i$ & $i^\mathrm{th}$ grid cell \\ 
 \hline
$t$ & Time Slots \\
\hline
$t_i$  & $i^\mathrm{th}$ time slot%Time between $(i-1)^\mathrm{th}$ and $i^\mathrm{th}$ hour 
 \\
 \hline
$R$ & Request sequence \\
\hline
$r_i$ & $i^\mathrm{th}$ request \\ \hline
$r_{s_i}$, $r_{d_i}$ & Source and destination of request $i$ \\ \hline
$r_i^t$ & Time at which request $i$ arrives \\ \hline
$G$ & Request Graph \\ \hline
OD Matrix & Adjacency matrix of $G$ \\ \hline
$V$ & Vertices of graph $G$ \\ \hline
$E$ & Edges of graph $G$ \\ \hline
$G_i$ & Graph $G$ at $t_i$ time slot \\ \hline
$\Delta_\mathrm{ij}$, $\hat{\Delta_\mathrm{ij}}$ & Actual and predicted number of requests between grid cells $g_i$ and $g_j$
\\ \hline
$\delta_\mathrm{i}$, $\hat{\delta_\mathrm{i}}$ & Actual and predicted number of requests at grid cell $g_i$ 
\\ \hline
$f_i^t$, $b_i^t$, $q_i$ & Set of forward, backward, and geographical neighbors of grid cell $g_i$ at time $t$\\ \hline
$e_i^t$ & Embedding of node $v_i$ at time $t$ \\
\hline
$W_c,W_s$ & Learnable weight matrices\\ \hline
$w'$ & Pre-weighted aggregator \\ \hline
$Y$  & Vector that concatenates the embedding of neighboring nodes \\ \hline
$a$ & Attention coefficient which maps a vector to a single number \\ \hline
$\mu$ & $LeakyReLu$ activation function \\ \hline
$f_\mathrm{ij}^t$, $b_\mathrm{ij}^t$,$q_\mathrm{ij}^t$ & Weight of forward, backward and geographical neighbors when  embeddings are exchanged \\ \hline
$\alpha_j^t,\beta_j^t,\gamma_j^t$ & Pre-weighted aggregator for forward, backward and geographical neighbors \\ \hline
$D$ & Graph that stores distance between different grid cells \\ \hline
$d_\mathrm{ij}$ & Distance between grid cells $g_i$ and $g_j$ \\ \hline
$h$ & Number of hours for which non-linearity is determined  \\ \hline
$W^K,W^Q,W^V$ & Learnable weight matrices called as key, value and query matrices respectively \\ \hline
$E^t,E_i^{t+1}$ & Spatial and initial embeddings of $n$ grid cells at time slot $t$ and $t+1$ respectively \\ \hline
$S$ & Similarity matrix   \\ \hline
$S',S''$ & Scaled and normalized similarity matrices \\ \hline
$p_\mathrm{ij}$ & Probability that requests are transferred from grid cell $g_i$ to grid cell $g_j$ \\ \hline
$m$ & Total  number of requests \\ \hline
$n$ & Total number of grids \\ \hline
\end{tabular}}
\caption{Notations}
\label{table:notations}
\end{center}
\end{table}

%The graph generated from pre-processing stage is transformed into a form where it can processed by the GNN model. %a stand-alone graph and does not have any information  
Each vertex of the graph is represented by an \emph{embedding} which is its transformation into a vector space that describes it completely and preserves the maximal information about the local structure of the graph (connection between nodes and edges). In our proposed model the initial embedding of a vertex is  a combination of its grid ID ($g_i$ for the cell $i$ of grid), row number, column number, time slot $t_i$, day of the week, in-degree, and out-degree. The in-degree and out-degree of a node are calculated from the OD matrix. If we represent each element of the OD matrix  as $OD(g_i,g_j)$ where $g_i$ represents the row grid cell and $g_j$  represents the column grid cell as can be seen through Figure \ref{fig:mesh2}, then the in-degree  of the grid cell $g_k$ is  $\sum_{i=1}^n OD(g_i,g_k)$ and its out-degree is  $\sum_{j=1}^n OD(g_k,g_j)$. %For instance, the grid ID of the grid cell $g_1$ is 1, its in-degree is $\sum_i g_ig_1$
%In our proposed model the initial embedding of a vertex is  a combination of  its grid ID ($g_i$ for the cell $i$ of grid) and its in-degree, and out-degree that are calculated from the OD graph shown in Figure \ref{fig:mesh2}. 
Initially, the embeddings of a node contain the local information associated with the node and edge connections. This information %about the local neighborhood 
is unstable as it only provides a view of the local neighborhood and does not provide any indication of the spatial or temporal dependencies that can arise among the requests. %that arrive in ride-hailing platforms and is therefore unstable. %and it only reveals the structure of graph %are constructed by concatenating the information from the local neighborhood.
In order to capture the dependencies among the requests and obtain a global view of the network, the nodes exchange  embeddings with their spatial and temporal neighbors. %The importance of spatial and temporal neighbors for the exchange of embeddings, our proposed model uses two layers:  %spatial and temporal  %other nodes. The nodes with which embeddings are exchanged are determined through the two proposed layers:
%These neighbors are selected by the \emph{Spatial Attention Layer}  and the \emph{Temporal Attention Layer} of our proposed model.
These neighbors are selected by the following two layers of our proposed model: \emph{spatial attention layer} - which selects the spatial neighbors and the \emph{temporal attention layer} - which selects the temporal neighbors. 
%Afterward, the nodes exchange embeddings with each other in an effort to obtain a broader view of the environment and capture the dependencies among requests. 
%These embeddings are exchanged with spatial and temporal neighbors.
%To analyze the relationship between different neighbors and thereby the exchange of embeddings our proposed model uses two layers: \emph{Spatial Attention Layer} and \emph{Temporal Attention Layer}. %These GNN layers exchange embeddings through the \emph{Graph Attention Networks (GANs)} wherein the nodes combine their vectors based on their weights. 
After the final embeddings are calculated through these two layers, % which reflect the spatial and temporal dependencies and have a global view of network, 
we need to predict the demand ($\hat{\delta_i}$) that can arrive at the $i^\mathrm{th}$  grid cell and the number of requests ($\hat{\Delta_\mathrm{ij}}$) that may arrive between the $i^\mathrm{th}$ and $j^\mathrm{th}$ grid cells. Here $\hat{\delta_i}$ and $\hat{\Delta_\mathrm{ij}}$ correspond to the predicted demand at grid cell $i$ and the predicted number of requests between grid cells $g_i$ and $g_j$ respectively.
 Whereas $\delta_\mathrm{i}$ and $\Delta_\mathrm{ij}$ represent the actual demand at grid cell $i$ and the actual number of requests between grid cells $g_i$ and $g_j$ respectively.
We get the demand at a node and the number of requests between two nodes  by feeding  
the result of these two layers  to the \emph{transferring attention layer}.
The working of these layers is described in detail in the next subsections.

\subsection{Spatial Attention Layer}
After the initial embedding of nodes is created, it is fed as input to the spatial attention layer which produces a new embedding that carries information about all the spatial neighbors. These embeddings are created by exchanging data with three different types of spatial neighbors: % The spatial neighbors with which embeddings are exchanged include : %by exchanging embeddings with three different types of neighbors. These neighbors include 

\begin{figure}[h]
\vspace*{-37mm}
    \centering
   % \hspace*{-25mm}
    \includegraphics[width=0.8\textwidth]{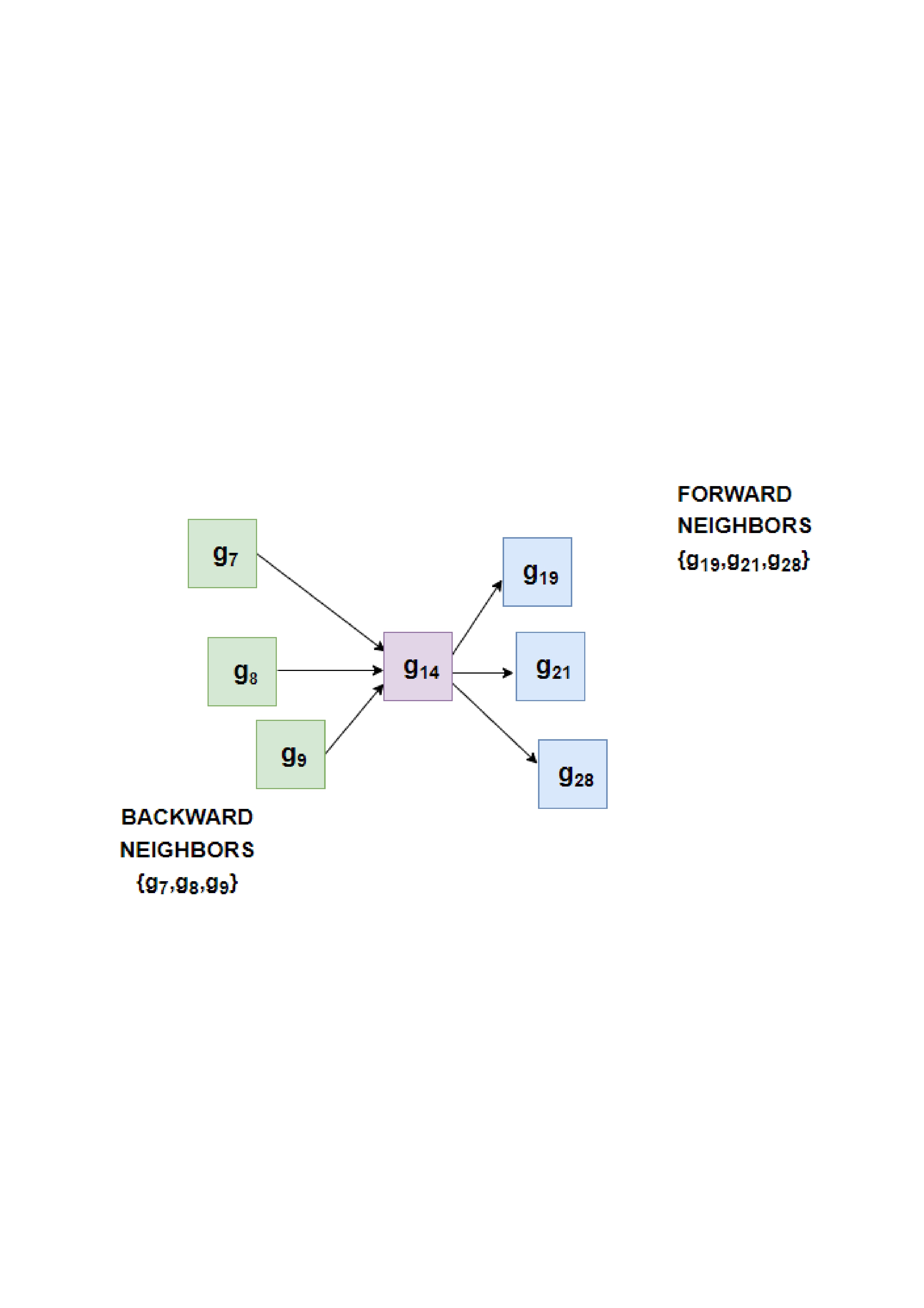}
    \vspace*{-38mm}
    \caption{Forward and backward neighbors calculated from an instance of graph $G$}
    \label{fig:fb}
\end{figure}
\textit{Forward neighbors} -  If there are two neighbors $g_i$ and $g_j$, and there is at least one request that originates from $g_i$ and is destined to $g_j$ ($\Delta_\mathrm{ij}>0$) then $g_j$ is the forward neighbor of $g_i$. The set of forward neighbors of $g_i$ at time slot $t$ is defined mathematically as:
\begin{equation}
    f^t_i=\{g_j \lvert\Delta_\mathrm{ij}^t>0, \Delta_\mathrm{ij}^t \, \epsilon \,G_t \}
\end{equation}
\textit{Backward neighbors} - If there are two neighbors $g_i$ and $g_j$, and there is at least one request that originates from $g_j$ and is destined to $g_i$ then $g_j$ is the backward neighbor of $g_i$. The set of backward neighbors of $g_i$ at time slot $t$ is defined mathematically as:
\begin{equation}
  b^t_i=\{g_j \lvert \Delta_\mathrm{ji}^t>0, \Delta_\mathrm{ji}^t \, \epsilon \,G_t \}  
\end{equation}
%$$$$
The sequential flow of requests in the network is captured by the forward and backward neighbors, which are also referred to as semantic neighbors. These neighbors identify passenger mobility patterns and determine the flow of requests into and out of the specific region. These neighbors are time-dependent, as the request flow in a region is not constant across different time slots. They are calculated from an instance of graph $G$. %through the OD matrix.  

Figure \ref{fig:fb} shows the set of forward and backward neighbors of the grid cell $g_\mathrm{14}$ at a particular time slot. The forward neighbors determine the out-flow of requests from a particular grid cell and the backward neighbors determine the in-flow of requests to a particular grid cell. 
As can be seen through Figure \ref{fig:fb},   $\{g_\mathrm{19}$,$g_\mathrm{21}$, $g_\mathrm{28}\}$ is the set of  forward neighbors of grid cell $g_\mathrm{14}$ as there are some requests that originate from $g_\mathrm{14}$ and are destined towards $g_\mathrm{19}$, $g_\mathrm{21}$ and $g_\mathrm{28}$ respectively. Similarly $\{g_7,g_8,g_9\}$  is the set of  backward neighbors of grid cell $g_\mathrm{14}$ as the requests from grid cells $g_\mathrm{7}$,$g_\mathrm{8}$ and $g_\mathrm{9}$ have their destination  at grid cell $g_\mathrm{14}$.  

\textit{Geographical neighbors} - Two neighbors $g_i$ and $g_j$ are said to be geographically connected if the Haversine distance between their corresponding latitude/longitude pairs is within a specified threshold distance. 
\begin{equation}
  q_i=\{g_j \lvert d_\mathrm{ij} \leq L, d_\mathrm{ij} \, \epsilon \, D \}  
\end{equation}
%$$$$ 
where $L$ is the threshold distance that determines the size of geographical neighbors and $D$ is the distance graph that represents the distance ($d_\mathrm{ij})$ between the central points of grid cells $g_i$ and $g_j$. 
For instance, if we consider the part of the road network represented in the form of a grid by Figure \ref{fig:mesh2}, where the threshold distance is set to be equal to the length of one grid cell, then the set of geographical neighbors of grid cell $g_\mathrm{14}$ are
is $\{g_8,g_9,g_\mathrm{10}, g_\mathrm{13}, g_\mathrm{15},g_\mathrm{18},g_\mathrm{19},g_\mathrm{20}\}$. The set of geographical neighbors of $g_i$ is constant across all time slots. % $t_k, \, k \epsilon \,[1,24]$.

The geographical neighbors can be used to aggregate uncertainty in information from the areas with few requests (sparse areas). As we know requests arrive in negligible quantity in sparse areas, and the forward and backward neighbors capture the dependencies between the neighbors based on the request flow of a particular region. % request flow of the region. %These neighbors capture the dependencies due to the in-flow and out-flow of requests to/from a particular region.
But, if a region is sparse in requests it will not have in-flow and out-flow of requests. In that case, there is no information from forward and backward neighbors. However, geographical neighbors are always there and can be used  to exchange embeddings in that area.  %and capture the flow of requests.   

%In sparse areas, the requests originate in minute quantity so there is hardly any flow of information from forward and backward neighbors. However, geographical neighbors are always there and can be used to extract knowledge about the areas. 

%After the initial embeddings of nodes are calculated, 
The initial embeddings of nodes are fed as input to the Graph Attention Network (GAT) which combines the information from the three neighbors described above and represents it in the form of a unified vector $e_i^t$ for each node $v_i$ at time $t$. Some of the earlier models have used Graph Convolution Network (GCN) \cite{Hamilton:neuips_2017} for combining information from different neighbors. However, with GCN all neighbors are assigned the same weightage when embeddings are merged, %and are  treated evenly 
which neglects the importance of nodes that have a similar flow of requests or that are  close to each other.
We propose to use GAT which samples different neighbors based on their weight. It assigns a higher value to the geographical nodes that are in the close vicinity of the current node. Similarly, it provides more weightage to the semantic neighbors with a higher flow of requests to/from the current node. In this way, embeddings of nodes that have more information are prioritized and the noise from redundant nodes is removed.
\begin{figure*}[t!]
%\vspace*{-25mm}
    \centering
   % \hspace*{-25mm}
    \includegraphics[width=1\textwidth]{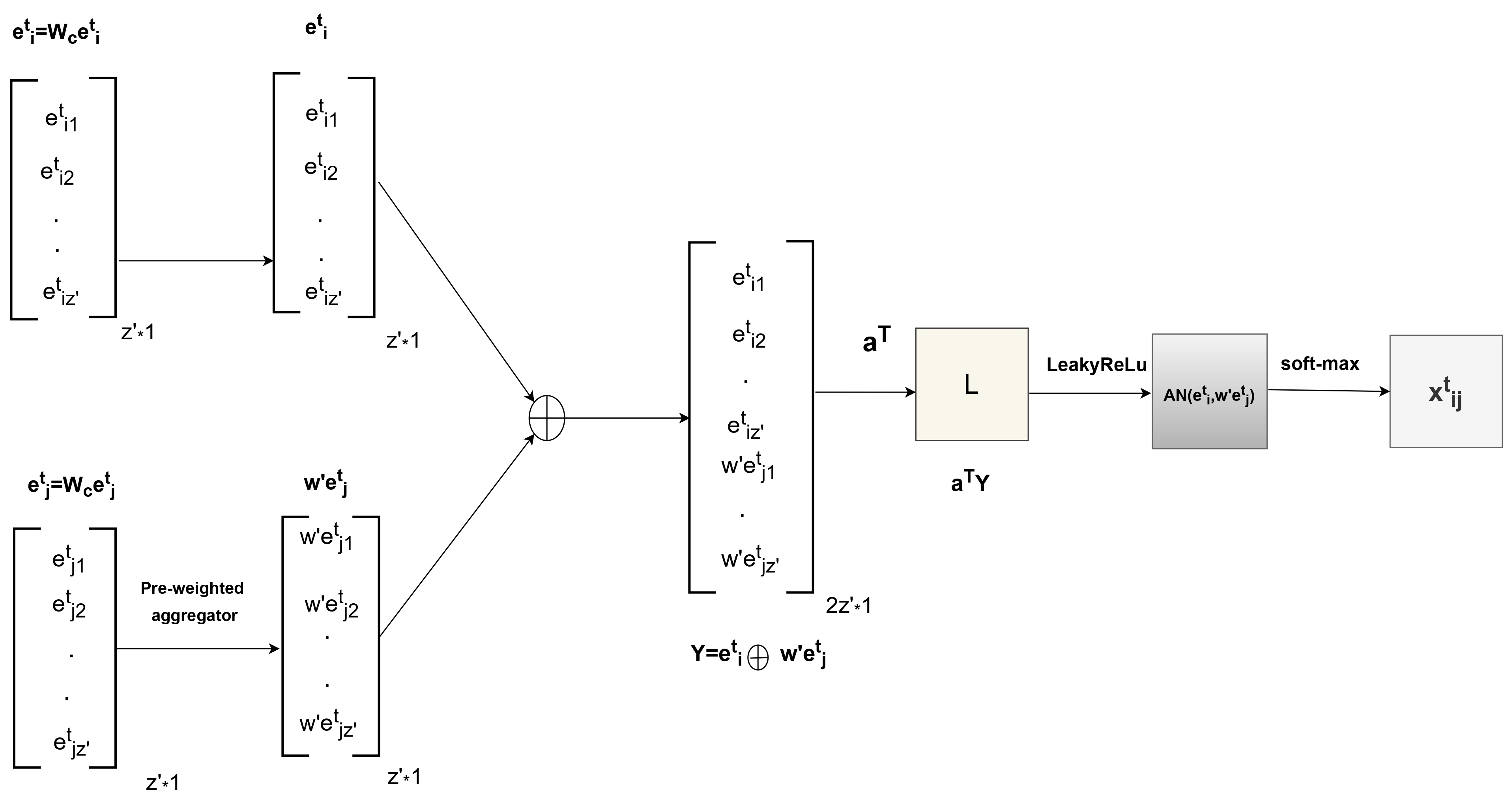}
  %  \vspace*{-8mm}
    \caption{Graph Attention Network for calculating affinity between neighboring nodes}
    \label{fig:an}
\end{figure*}

In order to calculate the importance of a neighborhood node $v_j$ at time $t$, we pass the  embedding of the current node $e_i^t \, \epsilon \, \mathcal{R}^\mathrm{z*1}$ and the neighborhood node $e_j^t \, \epsilon \, \mathcal{R}^\mathrm{z*1}$ through a weight matrix $W_c \, \epsilon \, \mathcal{R}^\mathrm{ z^{'}*z}$, where $z'>z$. This weight matrix acts as a single-layer neural network and projects the embedding of a node to a higher dimension. The output of this layer is the updated embedding of the nodes $v_i$and $v_j$ at time $t$ and they are equal to 
\begin{equation}
    e_i^t=W_ce_i^t 
\end{equation}
\begin{equation}
    e_j^t=W_ce_j^t 
\end{equation}
where $e_i^t \, \epsilon \, \mathcal{R}^\mathrm{z^{'}*1}$ and $e_j^t \, \epsilon \, \mathcal{R}^\mathrm{z^{'}*1}$. Figure \ref{fig:an} displays the working of the attention-based aggregator of GAT.
The input to GAT is the embedding $e_i^t$ and $e_j^t$ of nodes $v_i$ and $v_j$ at time $t$, after they have been passed through weight matrix $W_c$. The embedding $e_j^t$ is  passed through a pre-weighted aggregator $w^{'}$ which provides a prior weight to it before the GAT calculates its importance. This prior weight is based upon the current state of neighbors and is described in detail in the next subsection.  The embedding $e_i^t$ of node $i$ and the embedding of the neighborhood node after being passed through a pre-weighted aggregator ($w^{'}e_j^t$)
are concatenated and represented as a single vector $Y$:
\begin{equation}
    Y=e_i^t  
    \oplus w^{'} e_j^t
\end{equation}
%The embeddings are then passed through the 
This vector is then passed through a learnable attention coefficient $a \, \epsilon \, \mathcal{R}^\mathrm{2z^{'}*1}$ which maps it to a single number. Thereafter, non-linearity is applied through the $Leaky ReLu$ activation function denoted as $\mu$ in Eq.\eqref{eq:an}. 
%  and determines the importance, and non-linearity is applied to it through the $Leaky ReLu(\mu)$ activation function. .
 \begin{equation}
     AN(e_i^t,w'e_j^t)=\mu(a^T Y)
 \label{eq:an}
 \end{equation}
%and applies non-linearity to it by using $LeakyReLu(\mu)$ activation function. 
%These steps can be followed through the Figure \ref{fig:framework} which describes the working of attention based aggregator function $AN(v_i,w'v_j)$ of GAN described above . 
This attention-based aggregator function of GAT %$AN(e_i,w'e_j)$ 
measures the affinity between the embedding of node $v_i$ and its neighbor $v_j$ at time $t$ by learning the weight matrix $W_c$ and the attention coefficient $a$, and produces a single number that determines the weight of neighborhood node $v_j$ when it needs to exchange embedding with the node $v_i$.
%These steps merge the neighboring embeddings in a single vector and calculate the affinity between them by learning the weight matrix $W_c$ and attention coefficient $a$.

However, the output of the neural network is not normalized, which is a problem since the weights should be on the same scale for exchanging embeddings. 
In order to normalize the weights, we  apply the soft-max function to the output of the attention-based aggregator defined in Eq. \eqref{eq:an} which brings all the weights to the same scale. In Figure \ref{fig:an}, these normalized weights are denoted as $x_\mathrm{ij}^t$  and they represent the weights of forward neighbors ($f_\mathrm{ij}^t$), backward neighbors ($b_\mathrm{ij}^t$), and geographical neighbors ($g_\mathrm{ij}$). They are mathematically represented as %of every neighboring node as follows:
\begin{equation}
    \begin{split}
  f_\mathrm{ij}^t=\frac{exp(AN(e_i^t,\alpha_j^te_j^t))}{\sum_{k\,\epsilon f_i^t} exp(AN(e_i^t,\alpha_k^te_k^t))}\\      b_\mathrm{ij}^t=\frac{exp(AN(e_i^t,\beta_j^te_j^t))}{\sum_{k\,\epsilon b_i^t} exp(AN(e_i^t,\beta_k^te_k^t))}\\q_\mathrm{ij}^t=\frac{exp(AN(e_i^t,\gamma_je_j^t))}{\sum_{k\,\epsilon q_i^t}exp(AN(e_i^t,\gamma_ke_k^t))}
    \end{split}
\end{equation}

%$f_\mathrm{ij}^t$, $b_\mathrm{ij}^t$ and $q_\mathrm{ij}^t$ determine the weights of forward, backward and geographical neighbors ($v_j^t$)  when the node $v_i^t$ merges embedding with them. 
These weights allow our proposed model to prioritize embeddings that are geographically and semantically similar to the node $v_i$ at time $t$.  Here,  $\alpha_j^t$, $\beta_j^t$, and $\gamma_j$ refer to the pre-weighted functions which are detailed in the next subsection. % described above are described in the next subsection.%Here, an important point to note is that before the embeddings of node $v$ are fed into attention network they are weighted by factors $\alpha$, $\beta$ and $\gamma$ in equation \ref{}. These factors are called pre-weighted factors and their importance is described in next subsections.

 Since we have obtained the weights that determine the importance of each node, we exchange embeddings with the set of forward, backward, and geographical neighbors with the weights $f_\mathrm{ij}^t$, $b_\mathrm{ij}^t$ and $q_\mathrm{ij}^t$ calculated above as:%, and the sharable weight matrix $W_s$ as follows:
\begin{equation}
   e_i^t=W_s e_i^t \oplus \sum_{j \epsilon f_i^t} f_\mathrm{ij}^t  W_s e_j^t \oplus \sum_{j \epsilon b_i^t} b_\mathrm{ij}^t  W_s e_j^t \oplus \sum_{j \epsilon q_i} q_\mathrm{ij}^t W_s e_j^t 
\end{equation}
Here $e_i^t$ is the final embedding of each node $v_i$ at time $t$ and it is obtained by merging information from different sets of neighbors based upon their weights and $W_s$ is the shared weight matrix  that projects all embeddings onto the same $z^{'}$ dimensional space before merging them. Thereafter, they are passed through multi-head attention and head gates \cite{Shen_IEEEAccess:2022} in order to capture the relationship between grids.

\subsubsection{Pre-weighted aggregator}
The pre-weighted aggregator 
provide a prior weight to the embedding of node $v_j$ at time $t$ before its affinity is measured by the attention-based aggregator defined in Eq. \eqref{eq:an}.
The aggregator determines the current flow of requests, which is calculated from the edges ($\Delta_\mathrm{ij}$) of the graph $G_t$, and the geographical relationship between requests, which is determined through the edges ($d_\mathrm{ij}$) of graph $D$. It allows our proposed model to sense the sparsity or density of requests in the current time slot and accordingly aggregate the information from the forward and backward neighbors that have a higher flow of requests.  It also provides an indication of the geographical neighbors which are  closer to the requests that arrive at time $t$.
 These pre-weighted aggregators  are represented as $\alpha_j^t$, $\beta_j^t$ and $\gamma_j$ for the forward, backward and geographical neighbors respectively and are defined as: %in equation \ref{eq:1}. They are defined as:

\begin{equation}
    \begin{split}
  \alpha_j^t=\frac{\Delta_\mathrm{ij}}{\sum_{j \epsilon f_i^t}\Delta_\mathrm{ij} + s} \\      \beta_j^t=\frac{\Delta_\mathrm{ij}}{\sum_{j \epsilon b_i^t}\Delta_\mathrm{ij} + s} \\\gamma_j=\frac{\frac{1}{d_\mathrm{ij}}}{\sum_{j \epsilon q_i^t}\frac{1}{d_\mathrm{ij}}}
    \end{split}
\end{equation}
 $s$ is a small additive term to prevent the case when the denominator is equal to $0$.
 The weights $\alpha_j^t$ and $\beta_j^t$ represent the intensity of passenger demand at time $t$ and the weight $\gamma_j$ corresponds to the distance of request from node $i$. These weights help our proposed model to prioritize the embeddings that are geographically or semantically closer to node $v_i$ at time $t$. 

Table \ref{table:notations} provides a comprehensive summary of the notations used throughout the study. %It serves as a reference for understanding the symbols, abbreviations, and variables employed in the analysis

\subsection{Temporal Attention Layer}
After the data is filtered through the spatial attention layer, we get a low-level vector that carries information about all the neighborhood nodes and has an overview of the spatial dependencies of the graph. In order to capture the temporal dependencies among the learned representations, these node embeddings are exchanged with time-based neighbors using a temporal attention layer. The embeddings are exchanged through the scaled dot attention mechanism which will be described in detail next. This layer has $4$ channels: the first and second channels analyze the dependencies among requests that arrive in the %before and after the current hour 
preceding and consecutive hour of the previous $7$ days, and the third channel finds out dependencies among requests from the same hour of the previous $7$  days. These three channels capture the linear dependencies that arise due to the regular patterns in data like the morning and evening rush hours. %This layer exchanges the embedding through the same attention-based mechanism %described for 
%of the spatial attanetion layer.  
However, there could be non-linear dependencies which show up due to the recent events that might have taken place at the preceding time intervals or that reveal the travelling patterns of people. %denote the sparsity or density of requests in the area. %, or the travelling pattern of people. 
These dependencies are monitored by taking  the data from previous $h$ hours, where the value of $h$ is determined through experimental evaluations.  %They capture the flow of passengers in that area or monitor the travelling behavior of people. 
Their value provides an indication of customer flow in the region around the time request arrives and thereby helps our proposed model to capture the patterns apart from the morning and evening hour data repetition trends that are found in the existing studies. 

%The embeddings of different grid cells at a particular time slot $t$ are represented as $E_t=[e_1^t,e_2^t,....,e_n^t]^\intercal$, where $e_i^t$represents the embedding at grid cell $i$ at time $t$, and it captures the spatial dependencies in data. In order to capture temporal dependencies the proposed model exchanges these embeddings with $4$ layers. The embeddings of grid cells in these layers are represented as:
%\begin{equation}
 %   L_p=\{E_t\, | \, t=T-lp+1,\, p \,\in \, [1,P]\}
%\end{equation}

%\begin{equation}
 %   L_n=\{E_t\, | \, t=T-lp,\, p \,\in \, [1,P]\}
%\end{equation}

%\begin{equation}
 %   L_s=\{E_t\, | \, t=T-lp+2,\, p \,\in \, [1,P]\}
%\end{equation}

%\begin{equation}
 %   L_{nl}=\{E_t\}_{t=T}^{T+h}
%\end{equation}

%In these equations, $L_s$ represent the embeddings of the grid cells calculated at the same hour for the previous $P$ days,  $L_p$, and $L_n$, represent the embeddings calculated at the subsequent hours $T-1$ and $T+1$ for the previous $P$ days, and $L_{nl}$ represents the data of previous $h$ hours. Thus, each layer contains the embeddings of different time slots which are either from the previous $P$ days for linear data, or the current day for non-linear data. 

\begin{figure*}[t!]
%\vspace*{-25mm}
    \centering
   % \hspace*{-25mm}
    \includegraphics[width=1\textwidth]{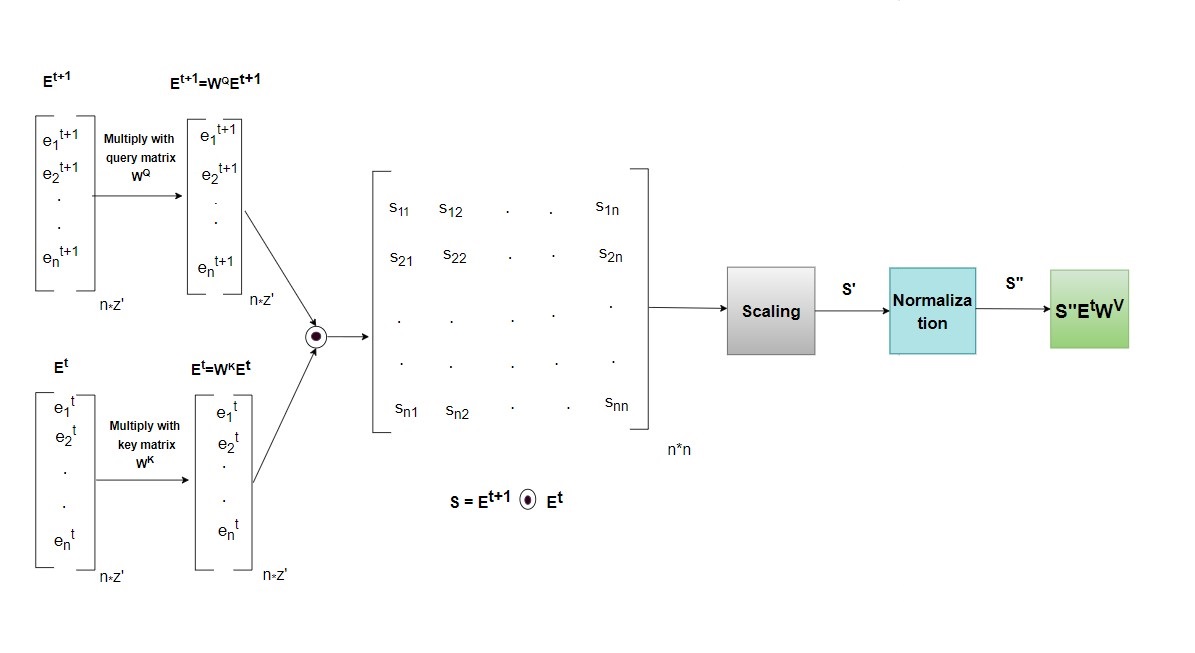}
  %  \vspace*{-8mm}
    \caption{Scaled dot attention for calculating affinity between temporal embeddings}
    \label{fig:scaled_dot}
\end{figure*}

The temporal attention layer determines the dependencies between the embeddings of different time slots through the use of the scaled dot attention mechanism which has been found to capture the dependencies among graphs in a better manner than the self-attention mechanism \cite{Wang:ACMTrans_2022}. 
Figure \ref{fig:scaled_dot} illustrates the mechanism followed by the scaled dot attention layer in determining the importance of embedding from a particular time slot. 
It gets as input the initial embeddings $E^{t+1}$  of all the grid cells at the time slot $t+1$ when prediction needs to be done, and the embeddings $E^t$ of the previous time slot $t$ whose weightage or importance needs to be determined. These embeddings are passed through the learnable weight matrices $W^Q$ and $W^K$ called the query and key matrices respectively. These learnable matrices act as a neural network and provide a better representation of the embeddings.  
The similarity between the embedding $E^{t+1}$ and $E^t$ is measured through the use of a dot product, which produces a similarity matrix $S$ whose elements $s_{ij}$ measure similarity between the $i^{th}$ grid cell at time slot $t+1$ and $j^{th}$ grid cell at time slot $t$.  
A high dot product value suggests dissimilarity between the embeddings, while a low value indicates similarity. 
 However, the dot product can produce values across a wide range, potentially leading to some values that are excessively large, consequently causing vanishing gradient issue. To mitigate this, scaling is applied to reduce the magnitude of these weights
After scaling, the value of this weight comes out to be any number between $-\infty$ to $\infty$ whereas the weights in the attention mechanism are between 
$0$ and $1$. Therefore, these weights are normalized using the soft-max function, resulting in the values in the desired range of  $0$ to $1$. %This illustrates the procedure for calculating the weightage or importance of embeddings at a particular time slot. % is how the weights are calculated for a particular time slot.
These weights measure the affinity between the grid cells at time $t+1$ with all the grid cells at time $t$. They are thereupon multiplied with the projected embeddings $E^t$ passed through the learnable weight matrix $W^V$, called the value matrix. Through this approach, the proposed system obtains an updated representation for each time slot $t$ by selectively focusing on grid cells that are similar to it.
The above procedure calculated the importance of embeddings at a particular time slot. However, the temporal dependency arises due to the $3$ linear layers and  $1$ non-linear layer, and there are multiple time slots in one layer. For instance, the non-linear layer contains the spatial embeddings of previous $h$ hours, and the linear layers contain the
spatial embeddings of previous $7$ days.
Thus for each layer, the proposed model sums the representations of all the time slots 
in order to obtain the temporal representation of the entire particular layer. Upon acquiring the embeddings for each layer, the proposed model consolidates this information into a unified spatio-temporal representation by incorporating another self-attention unit across all four layers.

After describing the procedure for determining the importance of temporal embeddings we describe the non-linear dependencies  that arise due to the contextual data and travelling patterns of people in detail next. 
 
 \subsubsection{Context-aware data}

Context-aware data refers to the data about the surrounding environment of a location and it provides an indication of the events that might have taken place around that location. This data is collected through various software and hardware tools and it has been used to modify the behavior of various systems. In the case of ride-hailing platforms, context-aware data can be used to direct the routes to users, considering the local events that might have taken place around that place. For instance, if the context-aware data displays that there is road construction in the area, then the vehicles are recommended to divert their route and travel through a different path in order to ensure the passengers reach their destination on time. % will not be directed towards that area. % in order to protect them. 
Similarly, if the context-aware data displays that the area is free from traffic jams, then the vehicles are directed towards that route to provide the users with a hassle-free transportation service. % as it will provide them with hustle-free. 
Various works have been done that have used context-aware data to direct safe \cite{Galbrun:ElsevierInforSys_2016}, pleasant \cite{desouza:IEEEConf_2018}, or personalized routes \cite{Shan:IEEEConf_2018} to the users. %These works have recommended users to switch their paths if the  context-aware data displays an unfavorable incident such as a crime event or a traffic jam has occurred. 
These works have used this data to modify the behavior of the system and direct users towards the routes that take into account the local condition of the place. % as the main factor for  the recommendation of routes to users.  %This data provides an indication of the events that take place around the corresponding location.  
%These works have used  various software and hardware tools to collect context-aware data and thereby have recommended personalized routes to the users. % keeping in consideration the data about the place. %it has been used by various ride-hailing optimization algorithms. %These algorithms have used this data for recommending safe, pleasant or personalized routes to the users. These algorithms collect the data from the surrounding through various tools and use it to recommend routes to users. 

Our proposed model also uses context-aware data to capture the events that might have taken place around the location at a particular period of time. 
However, in our proposed model the data is not gathered from the software and hardware tools. Rather it is collected by the non-linear channel of the temporal attention layer. This layer captures all the customer requests that have arrived in that area for the preceding $h$ hours and uses that to analyze the local environment of that place. %aggregates information about the requests of  that area from the previous $h$ hours. 
The data from preceding $h$ hours provides an indication of one of the following things: sparsity or density of requests in the area or the occurrence of an event at that place. %This data can also be collected 
%by various route
%Context awareness refers to the ability of a system  to collect information about its environment and modify its behavior accordingly. These systems use various software and hardware tools to automatically collect and analyze data and direct responses to the users. In the case of ride-hailing platforms various works have been done that have used context-aware information to direct safe \cite{Galbrun:ElsevierInforSys_2016}, pleasant \cite{desouza:IEEEConf_2018}, or personalized routes \cite{Shan:IEEEConf_2018} to users. Apart from using expensive software and hardware, context-aware data in ride-hailing platforms can be collected by aggregating information about the requests of  that area from the previous $h$ hours. This data from preceding $h$ hours provides an indication of one of the following things: sparsity or density of requests in the area or the occurrence of an event at that place. %This data can also be collected 

If the data for the past $h$ hours shows that there are few requests in the area, there may have been  some impulsive event that reduced the flow of requests in and out of the area. This event could correspond to heavy rainfall, snowfall, or a security breach in that area. %This data provides context-aware information about the place and can be used to provide safe or hustle-free routes to drivers. 
If there are no events
scheduled at the place and the customer demand is still low
then the area may have few requests, to begin with, meaning
that people in that area do not use ride-hailing platforms
for their daily commute. In both cases, there is a high probability that the area will have few requests in the next hour. This context-aware data thus helps our proposed model to anticipate future requests according to the local conditions of the place.  %might be sparse in requests from the very beginning i.e people in that area do not use ride-hailing platforms for their day to day travelling.
 %If that's not the case, 
 %If there are no events scheduled at the place and the customer demand is still low then the area may have few requests, to begin with, meaning that  people in that area  do not use ride-hailing platforms for their daily commute. This data reveals that the area is sparse in requests and it will not fetch more requests in near future. This data can be used to allocate the vehicles to the area according to its customer base.  
Similarly, if the previous hours are abundant in requests,  there might be an upcoming event in that area - a football game, festival sale,  or any other event that influences passengers. %This data can be used to direct more vehicles to that place. 
However, if there are no events scheduled and the customer demand is high, it reveals that the area is dense in requests and there is always a higher flow of passengers in and out of that area. %Accordingly, that place can be accommodated higher vehicles to keep the passengers on board. %more vehicles can be allocated to that place.
These patterns reveal important contextual information and can be used to monitor the flow of passengers in the region. They help our proposed model to capture local events that might have taken place around the location where the request needs to be predicted. %Apart from that, this information can be used to allocate vehicles to the area, or provide safe routes to drivers. %, apart from capturing spatio-temporal dependencies.
These dependencies are captured by the non-linear channel of the temporal attention layer. 
%This pattern in the area where requests are going to arrive is captured by the non-linear channel of Temporal Attention Layer. This layer provides context-aware information and gives an indication of  the local events of that area.
%These dependencies are captured by the non-linear channel of temporal attention layer. 
%Moreover, these depndencies also tell about the passenger behavioral patterns and the time period after which requests tend to re-appear over different areas.

\subsubsection{Travelling Behavior}

Non-linear dependencies can also arise due to the travelling patterns of people. 
According to a behavioral study, people spend $5$ to $6$ hours outside of their homes during holidays \cite{diffey_britishjournal:2011}, which can include time spent in their garden, walking,  or travelling in cars.  Thus, 
%we can say that 
after this time-period people tend to return to their original location by using any mode of transportation.  %service of various transportation modes.  
This suggests that requests start to recur after this time and these patterns can be exploited to predict future requests. %Our proposed model exploits these patterns by an
Further, during weekdays people go to their homes after their office hours are complete. As data from various countries reveals that people spend on average $2-6$ hours in their workplace \cite{Article:workhours}  this pattern can be exploited to predict the travelling behavior of people during weekdays. %This data also reveals that the requests reappear after the office hours of people are complete. 
Moreover, there are other recurring patterns such as shopping \cite{Article:workhours}, etc, which can be used to predict the future occurrence of requests. 
Thus, the data from previous  hours provides an indication of travelling behavior of people and it reveals their office hours, recreation time, shopping hours, and other hobbies that can be used to predict future requests. This behavior is captured by the non-linear channel of the temporal attention layer and it provides insights into the recurring pattern of requests in ride-hailing platforms which can be used to predict the future occurrence of requests.

\subsection{Transferring Attention Layer}
After the data is modelled for spatial and temporal patterns, we pass it through the feed-forward neural network layer to get the number of requests at each node of a graph which represents the demand in the region. However, the proposed model fine-tunes the output like \cite{Shen_IEEEAccess:2022} with the output of the historical average to obtain better results through the use of a weighted aggregator. The resultant demand is represented as 
a matrix $\hat{\delta}=\{\hat{\delta_1},\hat{\delta_2},...,\hat{\delta_n}\}$. 
%where $\hat{\delta_i}$ represents the predicted demand at node $i$. 
In order to represent the origin and destination of requests, the transferring attention layer is used which models the transmission from one node to another based upon the transferring probability $p_\mathrm{ij}$. $p_\mathrm{ij}$ represents the  probability 
that requests are transferred from node $i$ to node $j$ and it is mathematically represented as %calculated using the soft-max function. It is mathematically represented as:
\begin{equation}
    p_\mathrm{ij}=\frac{exp(AN(e_i^T,e_j^T))}{\sum_{i=j}^n exp(AN(e_i^T,e_j^T))}
\end{equation}
%Thus demand ($\delta$) is represented as a matrix $\delta=\{\delta_1,\delta_2,...,\delta_l\}$ 
 OD pair ($\hat{\Delta_\mathrm{ij}}$) is calculated from demand by using transferring probability $p_\mathrm{ij}$ as follows:
\begin{equation}
    \hat{\Delta_\mathrm{ij}}=\hat{\delta}\cdot p_\mathrm{ij}  %od_\mathrm{ij} \, \epsilon \, G_\mathrm{T+1}\; , \, d_\mathrm{ij} \, \epsilon D_\mathrm{T+1}  
    %\hat{m^2}
\end{equation} 

Like demand prediction, the output is fine-tuned  with the weighted aggregator \cite{Shen_IEEEAccess:2022} to obtain  better prediction results.

\section{Prototype Implementation}
\label{sec:implementation}

In this section, we evaluate our proposed model based on extensive simulations.

\textit{Datasets}

\begin{table}[h!]
    \centering
        \caption{Summary of Datasets used}
    \begin{tabular}{|c|c|c|}
      \hline   Datasets& New York & Washington DC  \\ \hline
        Time Span & 1 month & 1 month \\ \hline
        Grid cell size(default) & $2.5 \,km $ & $2.5  \,km$ \\ \hline
Number of grid cells & $361$ &$63$ \\ \hline
Time slot granularity & $1$ hour & $1$ hour \\ \hline
    \end{tabular}
    \label{tab:datasets}
    \end{table}

We conduct experiments to determine the parameter settings on the real-world taxi dataset generated from New York City. After that, we verify the parameter setting on the Washington DC dataset.  %The dataset is generated for New York City and contains data for the month of February.
%The performance of our proposed model is evaluated on two real-world datasets generated by Washington DC and New York. 
Table \ref{tab:datasets} provides a summary of the datasets used for the experimental evaluation of our proposed model. 
The dataset is divided into grid cells of lengths $2.3,2.4,2.5,2.6$ and $2.7\,km$  and the experiments are conducted on these sizes of grid cells. The default length of the grid cell is set as $2.5 \,km$ (it is discussed in detail in subsection \ref{sec:grid_length}). %The data is generated for the month of February $2016$. 
The datasets were collected for the months of February $2016$ and $2017$ respectively. Each dataset is divided into grid cells of $2.5\,km$ with a time gap of $1$ hour. The $1$ hour time interval is chosen because it corresponds to the movement patterns of people \cite{Wang:ACMTrans_2022}. For instance, rush hours typically refer to specific peak traffic times during the day, which occurs within this hourly framework. Moreover, using a 1-hour time slot enables the prediction model to perform one cycle of gradient descent and facilitates the optimizer in efficiently assigning vehicle requests.

The rows of the dataset are of the form pick-up time, pick-up latitude and longitude, drop-off latitude and longitude, and passenger count. This data
about passengers' origin and destination is fed as input to the GNN-based model and it predicts the number of requests that can arrive between any two locations within the next $1$ hour. %The predicted data contains the expected number of passengers  between any two locations or grid cells and is fed as input to the request graph $Q$ which is used for recommending routes to drivers.

\textit{Experimental Settings}

In the experimental work, we used $75\%$ of data for training purposes and kept the remaining $25\%$ for testing purposes. In the training settings, $10\%$ of data is used as a validation dataset and is used for hyperparameter testing. We implement our model with PyTorch $1.11.0$ on Python $3.9$. All the simulations were run on Windows i7 with 16 GB RAM for $200$ epochs. The model was trained with a batch size of $1$ and a learning rate of $0.001$. Based on these parameters, we evaluate the working of our proposed model.

\textit{Evaluation metrics and Loss function}

We evaluate the prediction accuracy of our proposed model based upon the two widely applied metrics: Mean Absolute Percentage Error (MAPE) and Mean Absolute Error (MAE) which are defined as:
\begin{equation}
  MAPE= \frac{1}{m} \sum_{i=1}^m \bigg \lvert \frac{\hat{y_i}-y_i}{y_i+1}  \bigg \rvert   
\end{equation}
\begin{equation}
    MAE=\frac{1}{m}  \sum_{i=1}^m \big \lvert \hat{y_i}-y_i \big \rvert
\end{equation}
where $m$ denotes the number of examples, $\hat{y_i}$ denotes the predicted result and $y_i$ represents the actual result. We estimate the performance of our proposed model on areas with varying levels of customer base and accordingly calculate MAPE-$0$, MAPE-$3$, and MAPE-$5$ where $0$, $3$, and $5$ denote the minimum number of requests in different areas. We have used this threshold to see the patterns in different areas according to their customer base. %The loss function used by our proposed model is SmoothL1Loss which is a variant of the mean-squared-error loss that uses a squared term if the absolute element-wise error falls below 1 and and L1 term otherwise.

The loss function used by our proposed model is \emph{SmoothL1Loss} \cite{Wang:ACMTrans_2022}, which is a variant of the mean-squared-error loss that employs a squared term when the absolute element-wise error falls below $1$ and L1 loss otherwise
It provides a smooth transition in the loss computation and ensures a balanced impact on the model's training. By adapting to the magnitude of the errors, the loss function effectively handles both small and large deviations, enhancing the model's overall performance and accuracy. This allows for more stable and reliable convergence during the optimization process.

\subsection{Grid cell length }
\label{sec:grid_length}
\begin{table}[t!]
\begin{center}
\caption{Simulation to check the length of a grid cell}
\label{table:grid}
\resizebox{\textwidth}{!}{  
\begin{tabular}{ | m{5em} | m{0.99cm}| m{1.3cm} | m{1.3cm}| m{1.3cm} || m{1.4cm}| m{1.4cm} | m{1.4cm}| } 
  \hline
 Task & \makecell{Grid\\ cell\\ length \\ (km)} & MAPE-0 & MAPE-3 & MAPE-5 & MAE-0 & MAE-3 & MAE-5 \\
  \hline
  OD & 2.3 2.4 2.5 2.6 2.7   & 0.4191  0.3913  \textbf{0.3866} 0.3876 0.3985  & 0.4036 0.3646 \textbf{0.3488} 0.3513  0.3850  &  0.3838 0.3466 \textbf{0.3269} 0.3299 0.3715 & 5.8907 5.8756 \textbf{5.4249} 5.4508 6.4091  & 15.3091 15.3788 \textbf{14.0479} 14.1229  16.9501 &  19.3616 19.5304 \textbf{17.7843} 17.8809 21.5885 \\ 
  \hline 
  Demand & 2.3 2.4 2.5 2.6 2.7  &  0.4350 \textbf{0.4173} 0.4222  0.4237 0.4405 & 0.3880 0.3688 \textbf{0.3612}  0.3627 0.3918 &  0.3671  0.3466 \textbf{0.3354} 0.3362 0.3717 & 53.7656 60.7930 \textbf{46.7006}  46.7409 60.4972  &  94.1438 106.6110 \textbf{81.6304} 81.6977 106.0421 & 106.8488 121.0631 \textbf{92.5871}  92.6591 120.4107 \\ 
  \hline
\end{tabular}}
\end{center}
\end{table}

The length of a grid cell is an important parameter that decides the complexity and accuracy of the proposed model. If the length is large, then the origin and destination of the majority of requests will be concentrated within a single grid cell and the OD prediction will reduce to demand prediction. If it is narrow, the time complexity of the model increases.   Thus the length of a particular grid cell is an important parameter and needs to be determined. 

Table \ref{table:grid} displays the performance of the proposed model based upon the parameters of MAPE and MAE, with the increase in length of a grid cell. There is a rapid decrease in the error of the model when the length  increases from $2.3$ $km$ to $2.5$ $km$. After that, the error becomes stagnant  for the grid cells of  length $2.5$ $km$ and $2.6$ $km$. Thereafter, the error is found to increase again.
The main perspective behind this behavior can be  the change in data size and the number of neighbors on the varying lengths of grid cells.
When the length of grid cell is small, there are multiple cells with no requests, and the spatial dependencies are not represented accurately by this size. Moreover, with small length time complexity of the model is high. However, when the length of grid cell increases, dataset size decreases, and the number of different neighbors changes which brings about an increase in error. Further, with an increase in grid cell length, the OD task is found to converge to demand prediction.  Thus the optimal value of the length of grid cells  needs to be determined which reflects passenger mobility and performs effectively in a time-bound manner. 
Based on the experiments we have set the length  of the grid cells as $2.5$ $km$ for our proposed model as it performs well  under all the evaluation metrics.

In summary, the experimental evaluation determined the grid cell size based on the spatial dependencies among locations and it was found that the decrease in grid cell size does not always increase the prediction accuracy of the model which can be attributed to complex spatial dependencies among different locations.
%Based on the outcomes of different evaluation metrics, the grid cell size was set as $2.5$. 
It's worth noting that ride-hailing platforms define their grid cell size based on a range of performance optimization factors, tailoring it to specific applications. However, the value determined by our proposed model can be adopted by other models when accuracy in prediction is of paramount importance, even if it means removing some finer-grained road details in favor of overall performance enhancement. %Although the ride-hailing platforms determine the grid cell size based upon various performance optimization factors and its value is tailored for specific applications, the value set by our proposed model can be used by different models when the accuracy of prediction holds importance even as some finer-grained road details are subtly affected in favor of improved overall performance. }

\subsection{Non-linearities}
\begin{figure*}[t!]
    \centering
    \vspace*{-21mm}
    \subfloat[MAPE-0 for OD prediction]{%
        \includegraphics[width=0.25\textwidth]{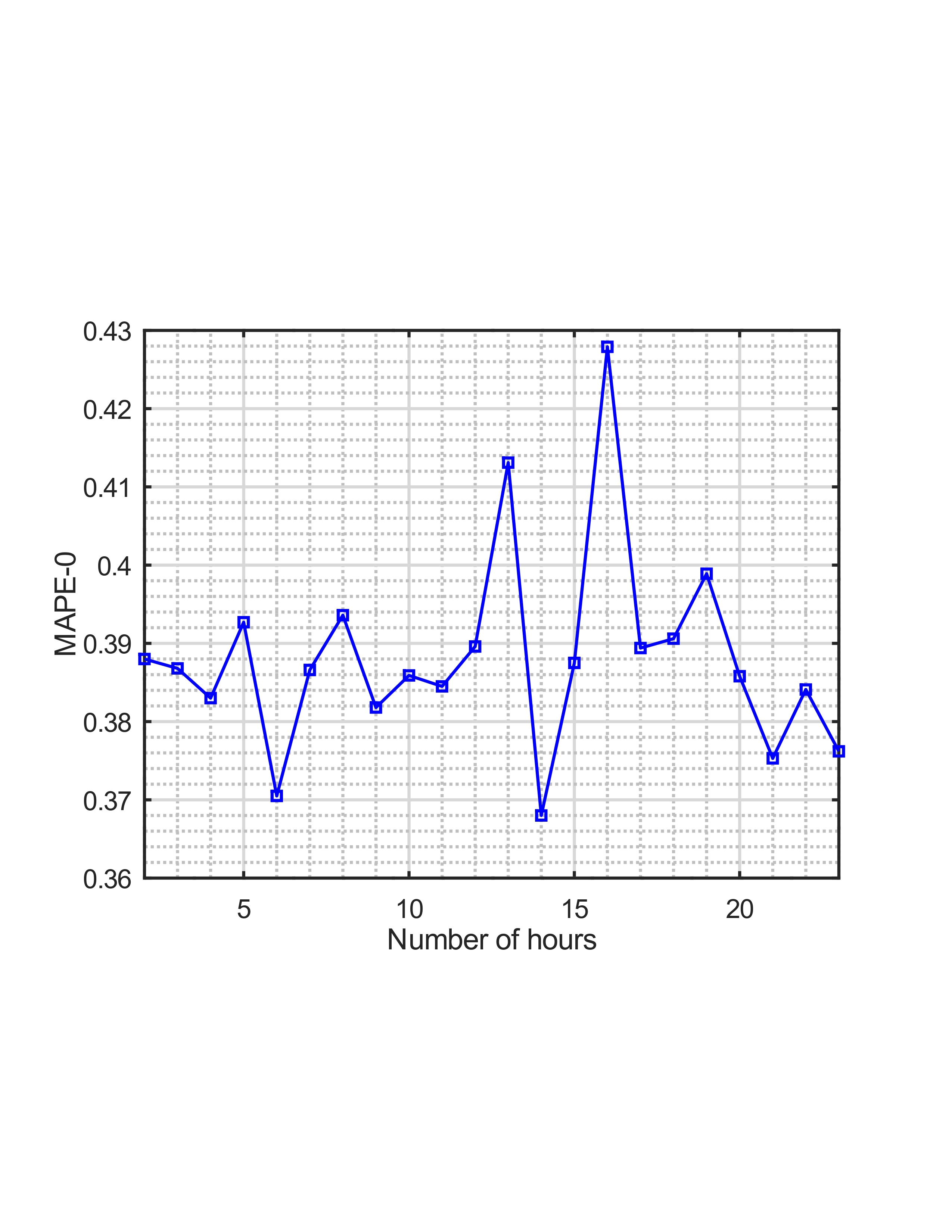}
        \label{fig:mape0od}
    }
    \hfill
    \subfloat[MAPE-3 for OD prediction]{%
        \includegraphics[width=0.25\textwidth]{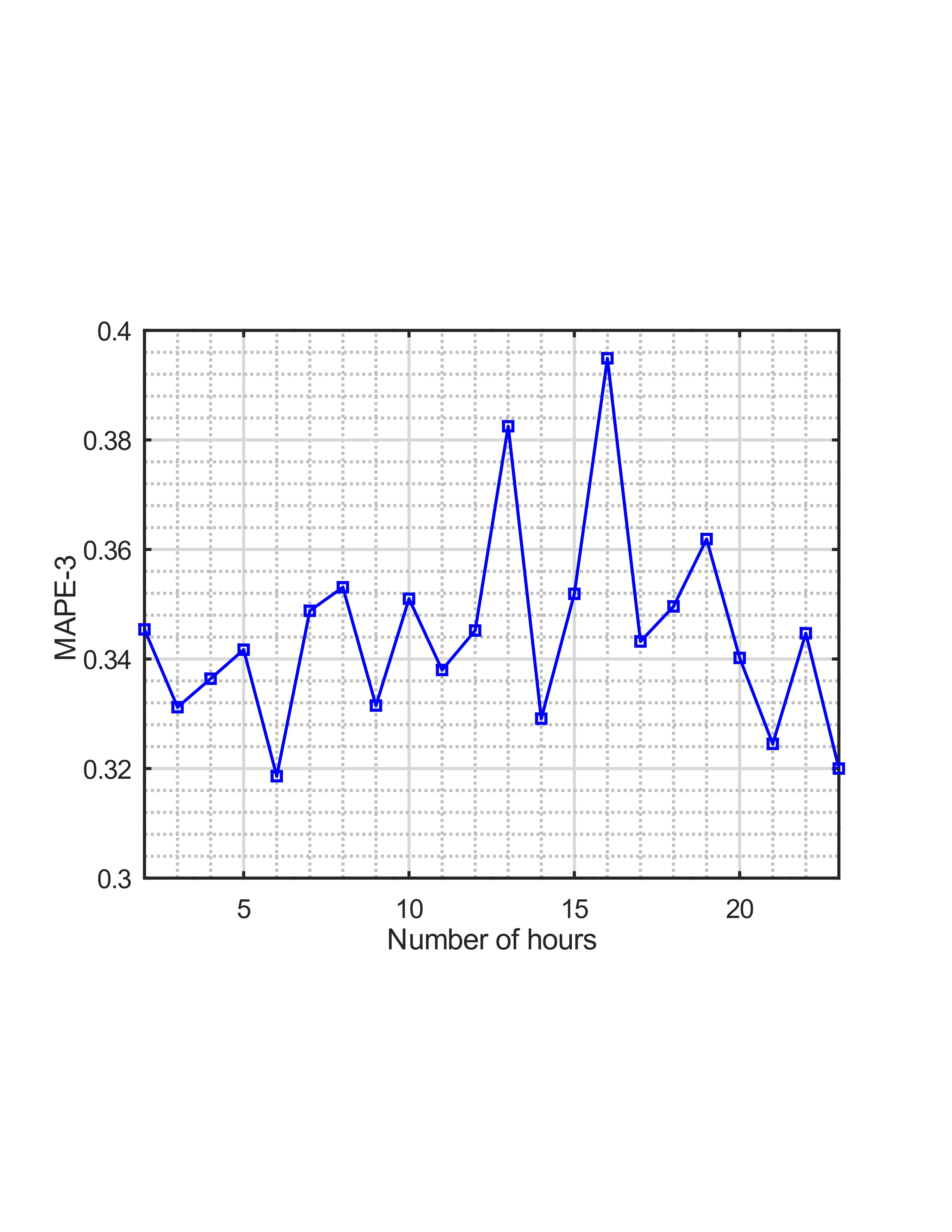}
        \label{fig:mape3od}
    }
    \hfill
    \subfloat[MAPE-5 for OD prediction]{%
        \includegraphics[width=0.25\textwidth]{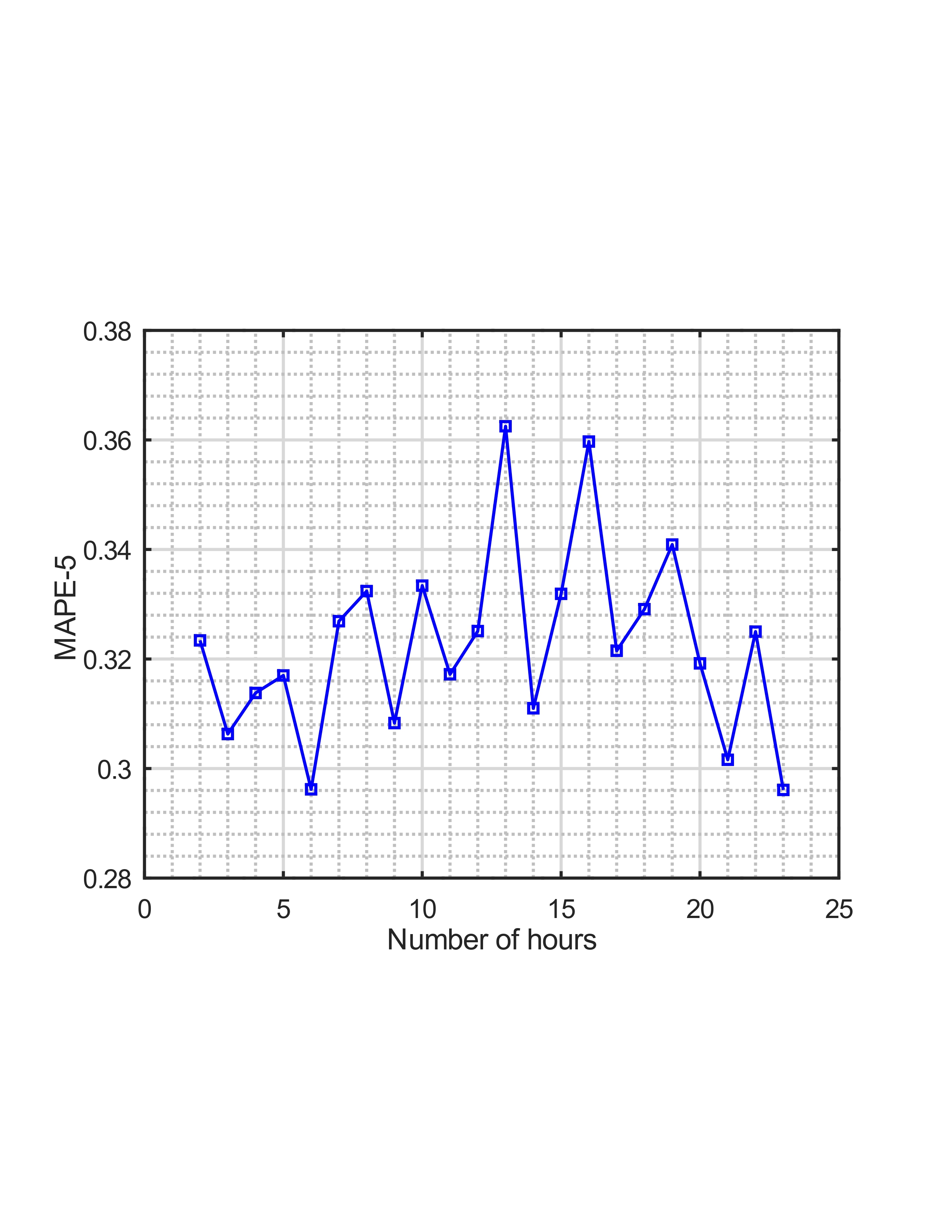}
        \label{fig:mape5od}
    }
    \caption{MAPE for OD prediction}
    \label{mapeod}
\end{figure*}

\begin{figure*}[t!]
    \vspace*{-8mm}
    \centering
    \subfloat[MAE-0 for OD prediction]{%
        \includegraphics[width=0.25\textwidth]{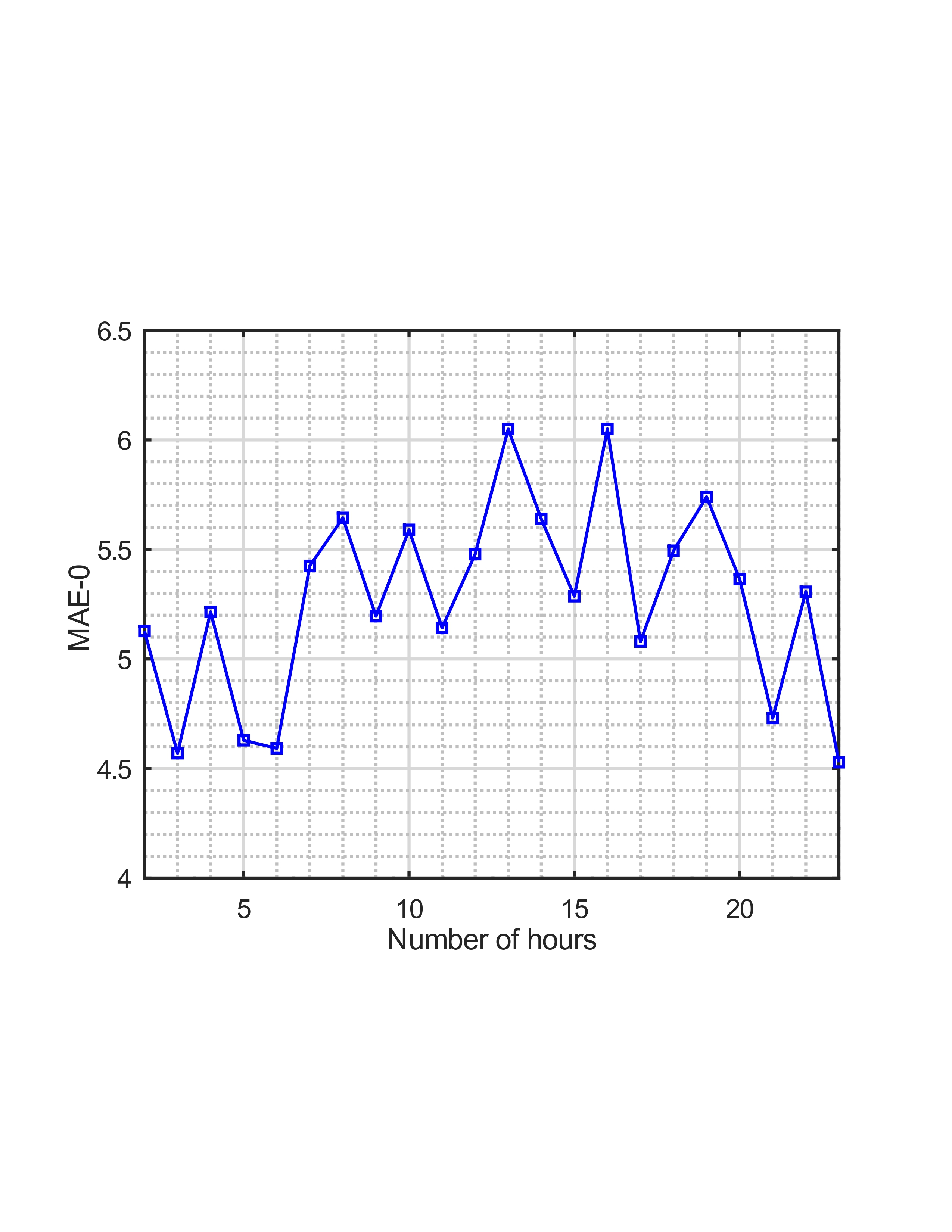}
        \label{fig:mae0od}
    }
    \hfill
    \subfloat[MAE-3 for OD prediction]{%
        \includegraphics[width=0.25\textwidth]{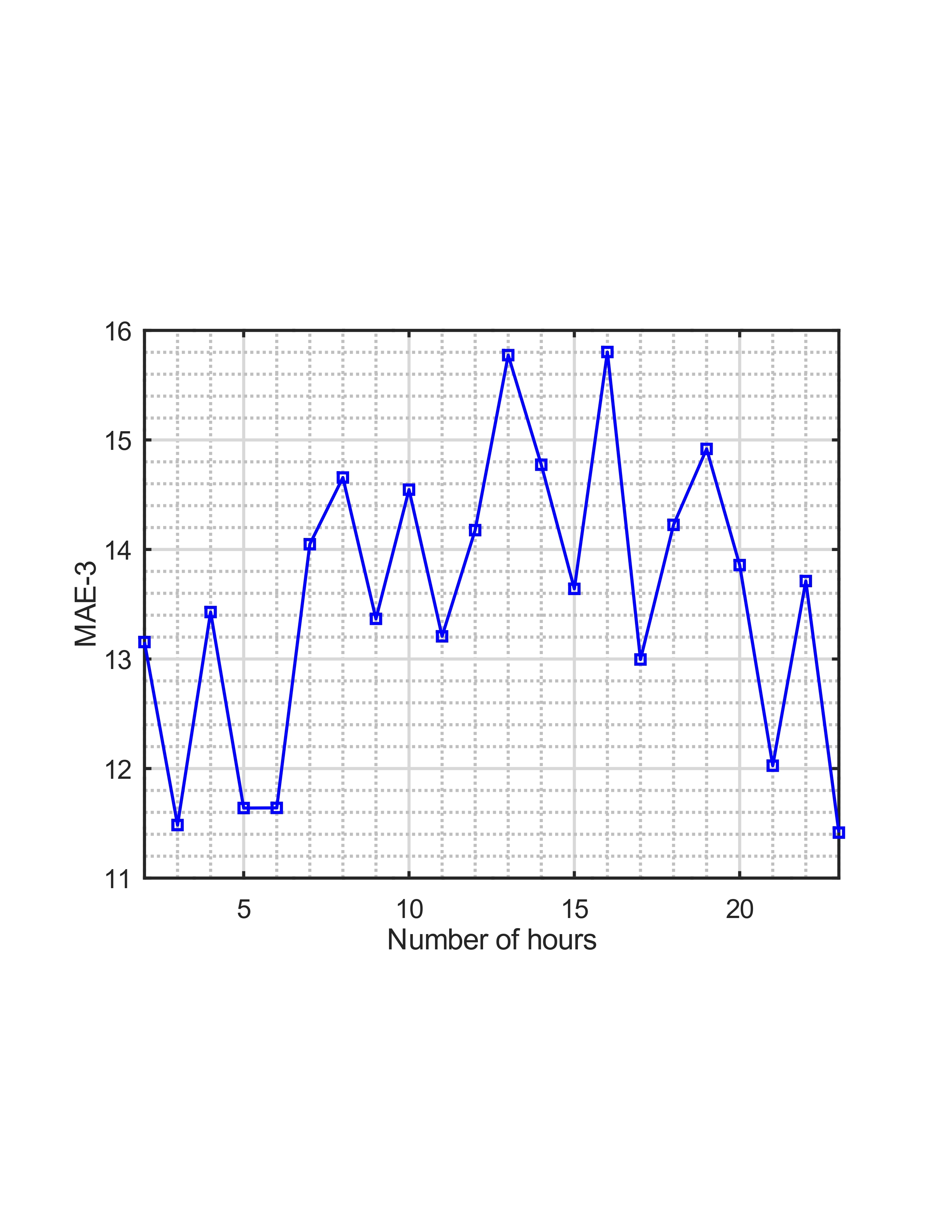}
        \label{fig:mae3od}
    }
    \hfill
    \subfloat[MAE-5 for OD prediction]{%
        \includegraphics[width=0.25\textwidth]{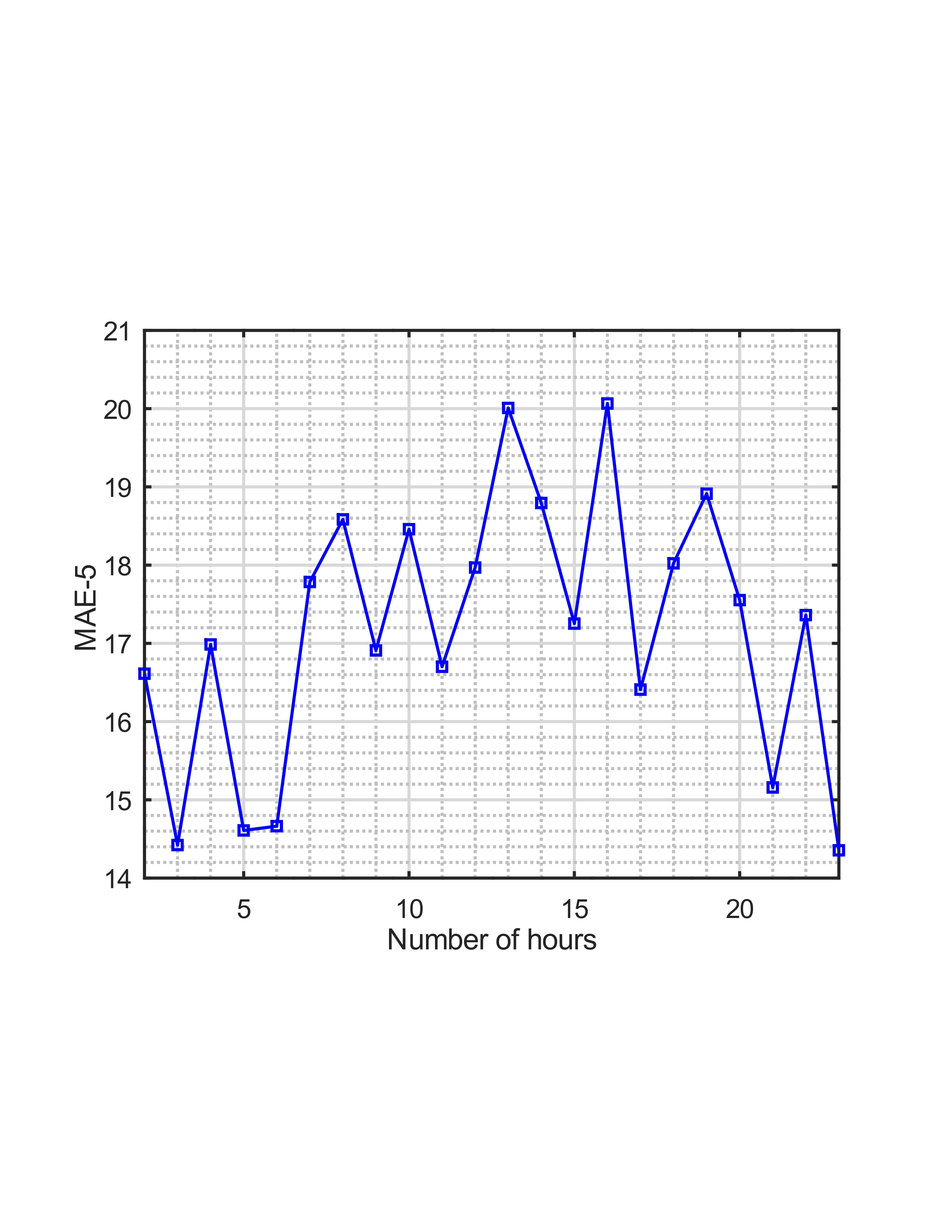}
        \label{fig:mae5od}
    }
    \caption{MAE for OD prediction}
    \label{maeod}
\end{figure*}

\begin{figure*}[t!]
    \vspace*{-8mm}
    \centering
    \subfloat[MAPE-0 for demand prediction]{%
        \includegraphics[width=0.25\textwidth]{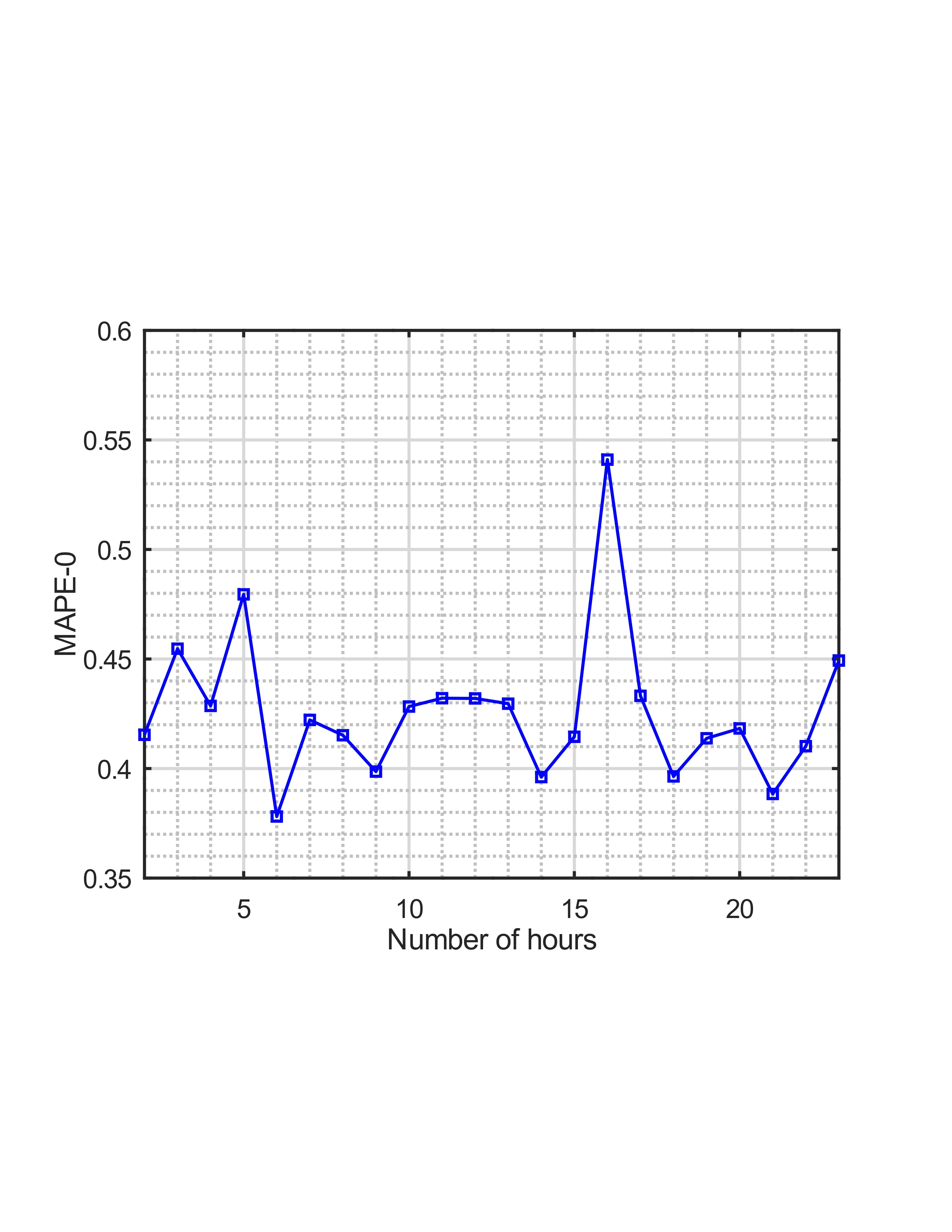}
        \label{fig:mape0demand}
    }
    \hfill
    \subfloat[MAPE-3 for demand prediction]{%
        \includegraphics[width=0.25\textwidth]{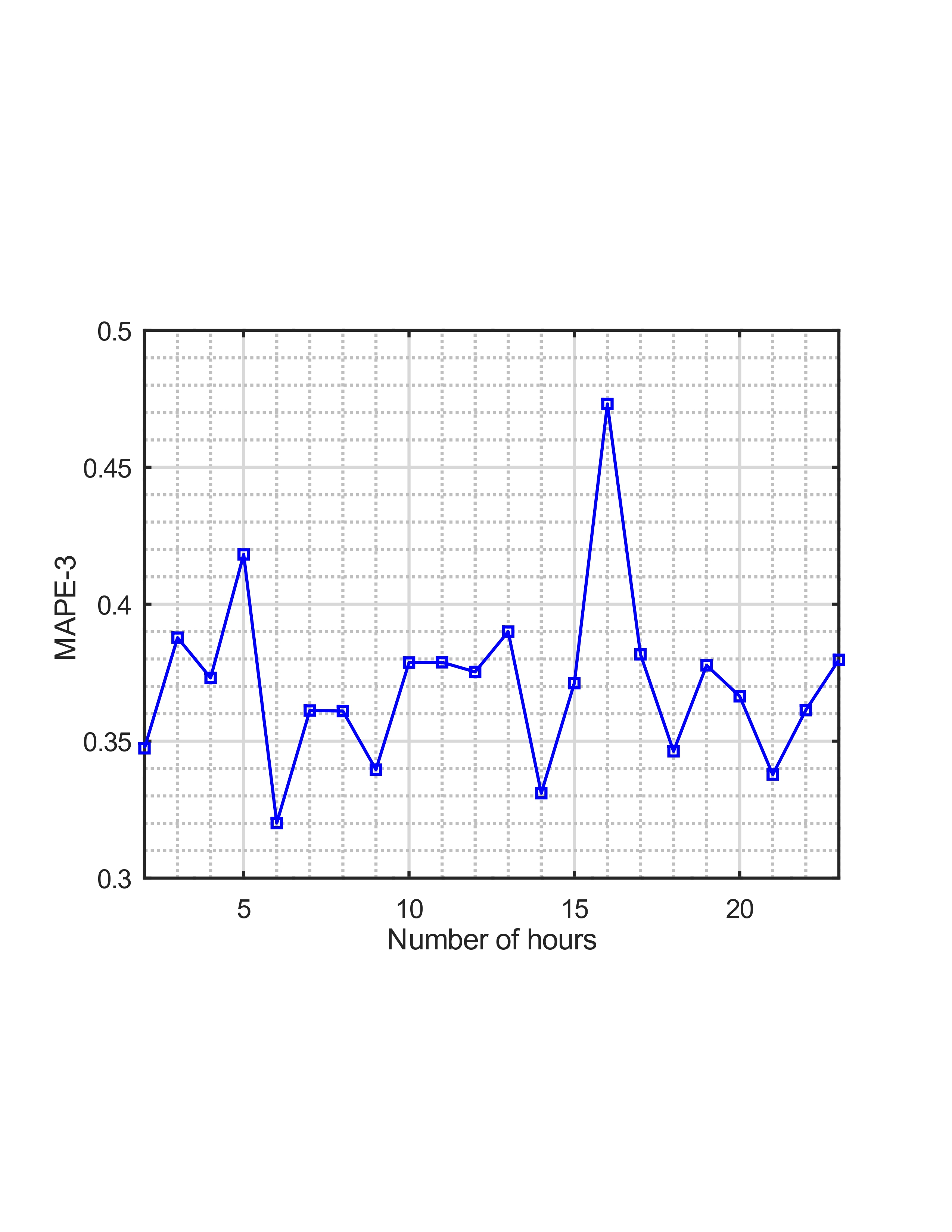}
        \label{fig:mape3demand}
    }
    \hfill
    \subfloat[MAPE-5 for demand prediction]{%
        \includegraphics[width=0.25\textwidth]{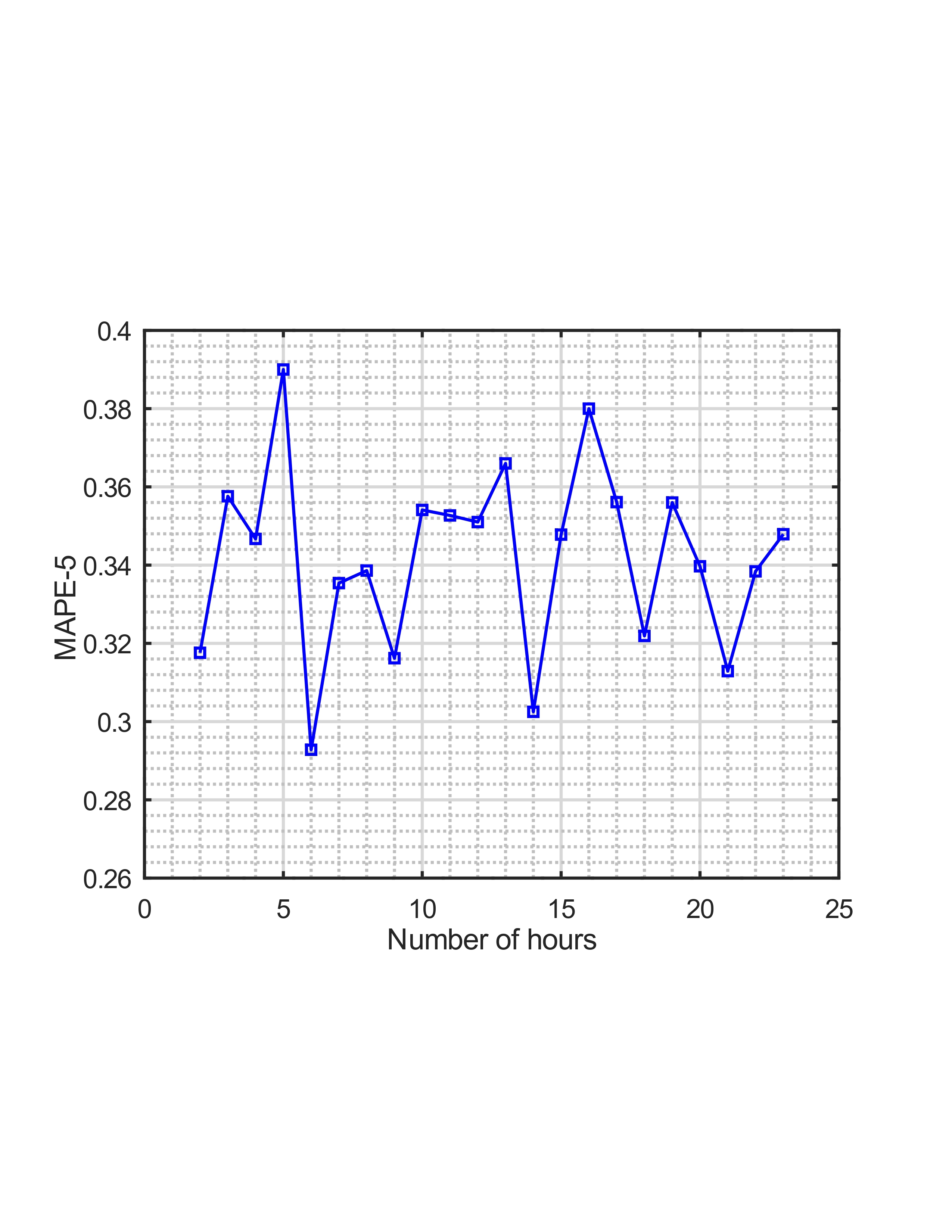}
        \label{fig:mape5demand}
    }
    \caption{MAPE for demand prediction}
    \label{mapedemand}
\end{figure*}

\begin{figure*}[h!]
    \vspace*{-21mm}
    \centering
    \subfloat[MAE-0 for demand prediction]{%
        \includegraphics[width=0.25\textwidth]{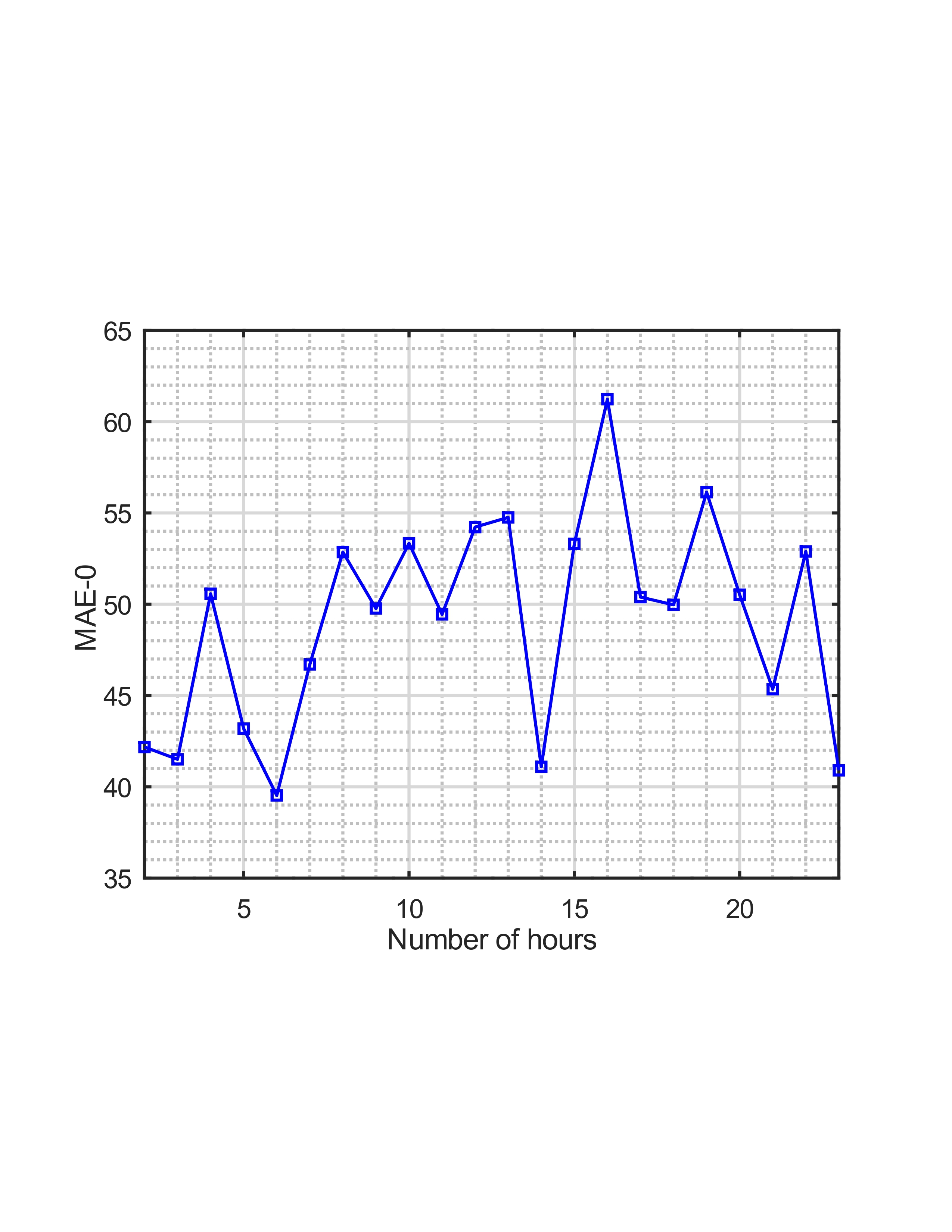}
        \label{fig:mae0demand}
    }
    \hfill
    \subfloat[MAE-3 for demand prediction]{%
        \includegraphics[width=0.25\textwidth]{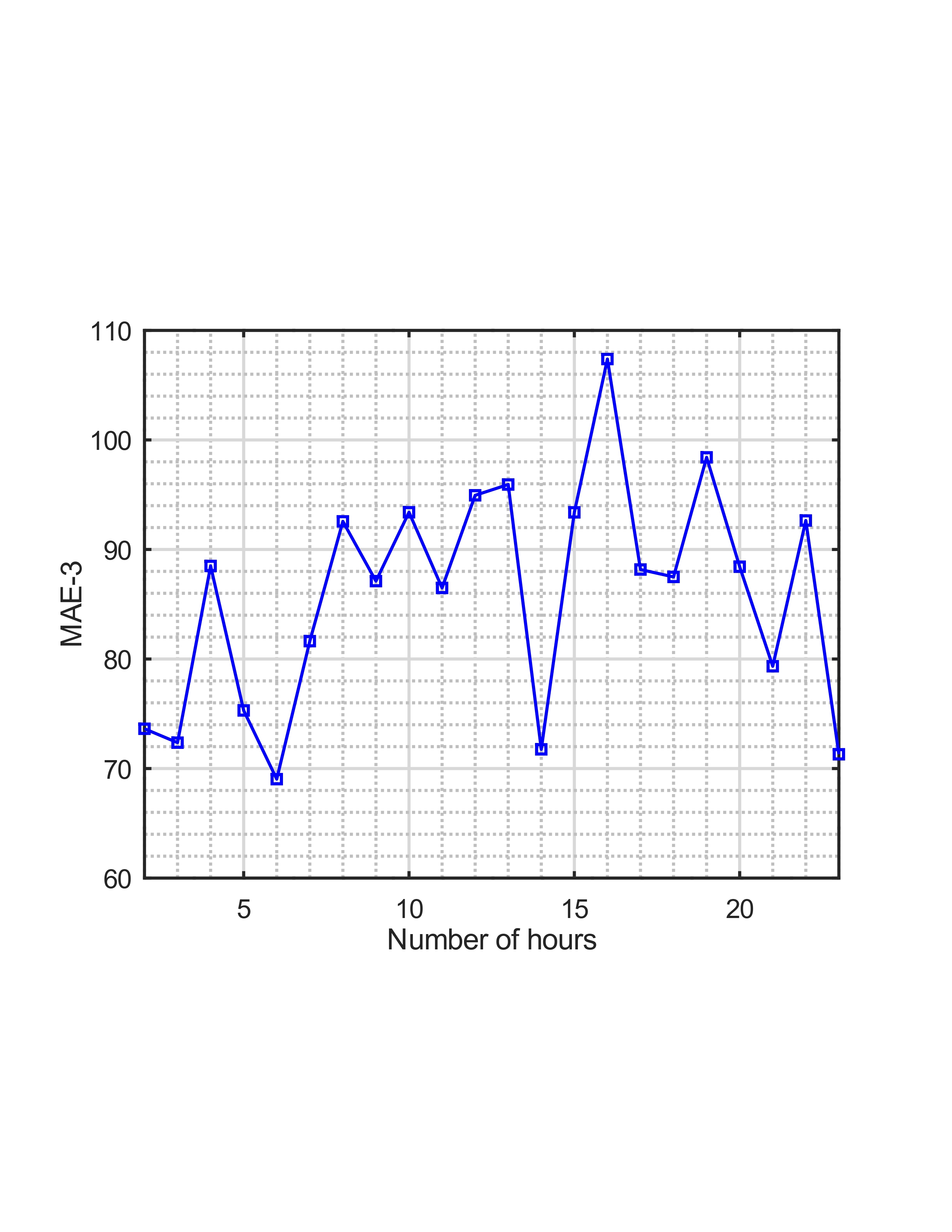}
        \label{fig:mae3demand}
    }
    \hfill
    \subfloat[MAE-5 for demand prediction]{%
    \vspace*{-704mm}
        \includegraphics[width=0.25\textwidth]{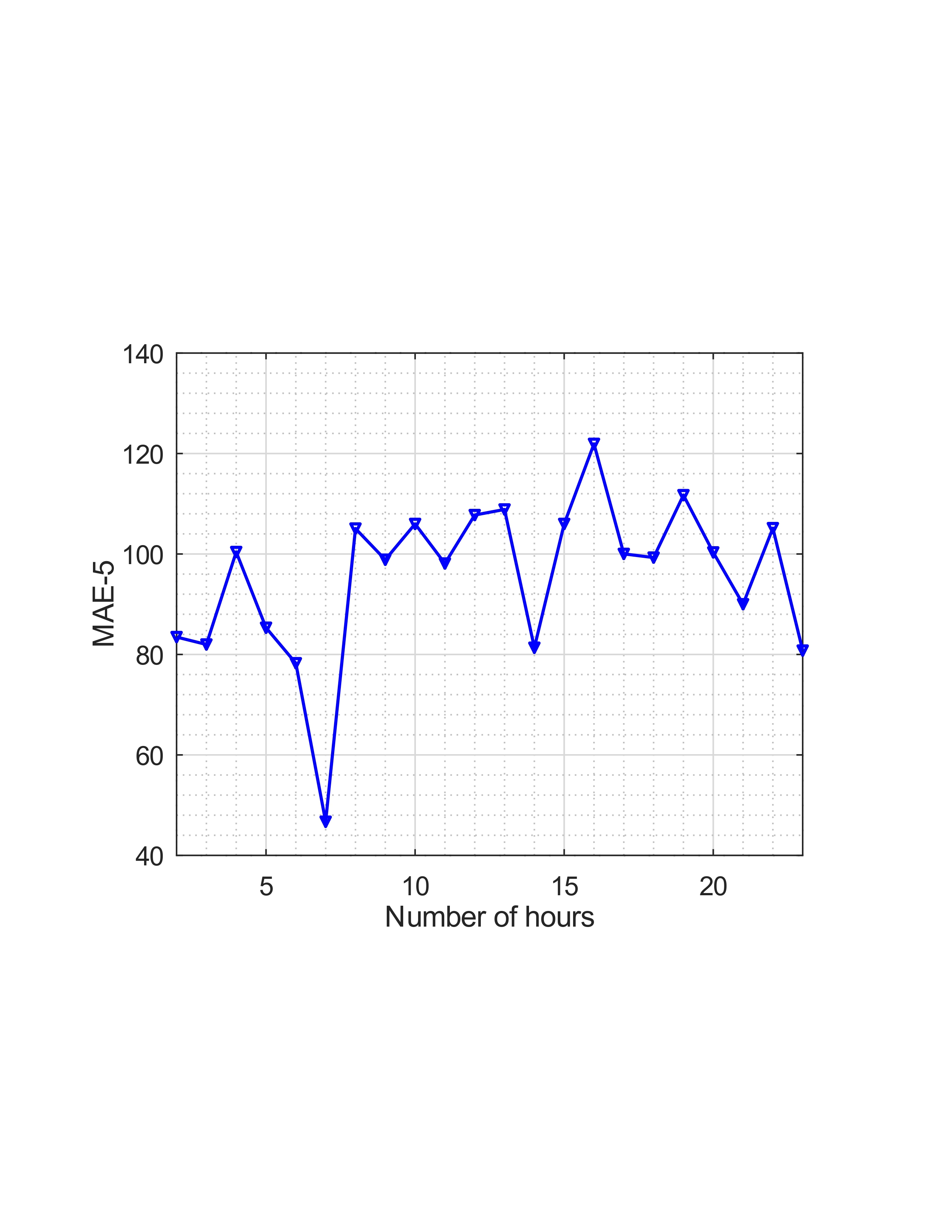}
        \label{fig:mae5demand}
  %      \vspace*{-100mm}
      %  \caption{MAE-5 for demand prediction}
    }
    \caption{MAE for demand prediction}
    \label{maedemand}
\end{figure*}

The non-linearities in our proposed model are captured by the fourth channel of the temporal attention layer. This layer monitors the data from previous $h$ hours and gives an indication of  
the events that might have taken place around the location during these hours. 
However, to properly monitor the events around that place, the value of $h$ needs to be determined through simulations. 
Figures \ref{mapeod}, \ref{maeod}, \ref{mapedemand} and \ref{maedemand} 
 show the MAPE and MAE of the proposed model for OD and demand prediction with varying time hours. As can be seen through these figures, when the data from the previous $3-6$ hours %and $20-23$ hours 
 is used, 
 the model is found to perform well under all the evaluation metrics. Particularly, the data from the previous $6$ hours is found to perform best under all conditions.

 As already stated, the non-linear data provides context-aware information about the location and determines the average flow of requests in the area over a period of time. For instance, if the data from the previous $3-6$ hours contains few requests then that area may be sparse or some event may have occurred like rainfall, security breach, etc which reduced the flow of requests to that area. Similarly, if the previous hours have more requests then that area will either be high-demand area or some event might be scheduled at that place which has increased its demand.

However, among the past $3-6$ hour data, the data from the previous $6$ hours is found to perform best under all the evaluation metrics. The intuitive explanation might be that it reveals the travelling behavior of people and represents the average time spent outside by users.

As the office timings in different countries vary from $2-6$ hours \cite{Article:workhours} and the demand re-arises after this time period, it %this 
provides one of the explanations for the data from the previous $6$ hours to perform well.
 Moreover, a behavioral study \cite{diffey_britishjournal:2011} has found that on average people spend $5-6$ hours outside their homes during holidays (it can include time spent in their garden, time doing walk or travelling in cars). It provides further explanation for using previous $6$ hour data to model non-linearity. %as the number of hours describing travelling behavior.
 Thus, from experimental studies and behavioral patterns of people, we conclude that data from previous $6$ hours reflects passenger mobility well and gives an indication of customer re-appearance over different areas.
% The data from $4-6$ hours represents the average time spent outside by users. As the office timings in different countries vary from $4-8$ hours and the demand re-arises after this time period, that is why data from $4-6$ hours is found to perform well. Moreover, various behavioral studies \cite{diffey_britishjournal:2011} have found that on average people spend $5-6$ hours outside their homes during holidays (it can include time spent in their garden, time doing walk or travelling in cars).
% Thus, $4-6$ hour data provides us with passenger behavioral patterns and gives an indication of customer re-appearance over different areas.

%Moreover, this data provides context-aware information about the location and determines the average flow of requests in the area over a time period. For instance, if the data from previous $6$ hours has fewer requests then that area may be sparse or some event may have occurred like rainfall, security breach etc. Similarly, if $6$ hour data has more requests then that area will either be high demand area or there might be some event like a match scheduled at that place which is captured through aggregate data from previous hours.

Data from previous $20-23$ hours is also found to perform well, in particular for OD prediction tasks as can be seen through Table \ref{table:repetitionpattern} and Figures \ref{mapeod} and \ref{maeod}. The main insight that we get from this pattern is the repeating sequence of requests the following day. As the $23$ hour data captures the average flow of requests for a day and this pattern repeats the following day, the origin and destination of requests are predicted accurately.

Thus we conclude that data from previous $6$ hours reflects the passenger travelling behavior and provides the contextual data of that place which can be used to analyze dependencies and predict the future demand and OD pair of requests in that area. Whereas the data from the previous $23$ hours captures the repeating trend in requests and uses that to predict the future OD pair. %After determining the parameters of non-linear layer, we describe the importance  of the proposed model

  \begin{table*}[ht!]
\begin{center}
\caption{Repetition pattern of requests}
\label{table:repetitionpattern}
\resizebox{\textwidth}{!}{  
\begin{tabular}{ | m{1.33cm} | m{0.5cm}| m{1.33cm} | m{1.33cm}| m{1.33cm} || m{1.5cm}| m{1.5cm} | m{1.5cm}| } 
  \hline
 Task & \makecell{Ho-\\urs} & MAPE-0 & MAPE-3 & MAPE-5 & MAE-0 & MAE-3 & MAE-5 \\
  \hline
  OD & 6 23  &  \textbf{0.3705}  0.3762 &  \textbf{0.3186} 0.3200   & 0.2962  \textbf{0.2961}   & 4.5922 \textbf{4.5285}   & 11.6408 \textbf{11.4151}  & 14.6635 \textbf{14.3549}  \\ 
  \hline
%  cell1 dummy text dummy text dummy text & cell5 & cell6 \\ 
  Demand & 6 23  & \textbf{0.3781}  0.4493  & \textbf{0.3201}  0.3797 &  \textbf{0.2928} 0.3479  &  \textbf{36.3080}  40.9023 & \textbf{61.8143} 71.3015   &  \textbf{70.1092}  80.7730\\ 
    \hline
\end{tabular}}
\end{center}
\end{table*}

\textit{Importance of non-linear layer}
  \begin{table*}[t!]
\begin{center}
\caption{Significance of non-linear layer}
\label{table:ablation}
\resizebox{\textwidth}{!}{  
\begin{tabular}{ | m{1.33cm} | m{0.8cm}| m{1.33cm} | m{1.33cm}| m{1.33cm} || m{1.5cm}| m{1.5cm} | m{1.5cm}| } 
  \hline
 Task & \makecell{Non-\\linear \\layer} & MAPE-0 & MAPE-3 & MAPE-5 & MAE-0 & MAE-3 & MAE-5 \\
  \hline
  OD & Yes No  &  \textbf{0.3705}  0.4791 &  \textbf{0.3186} 0.58655   & \textbf{0.2962}  0.59735   & \textbf{4.5922}  4.6253   & \textbf{11.6258} 11.6408  & 14.6635 \textbf{14.5918}  \\ 
  \hline
%  cell1 dummy text dummy text dummy text & cell5 & cell6 \\ 
Demand & Yes No  & \textbf{0.3781}  0.43515  & \textbf{0.3201}  0.4048 &  \textbf{0.2928} 0.387 &  \textbf{36.3080}  38.7564  & \textbf{61.8143} 67.5454   & \textbf{70.1092}  76.5284\\ 
 \hline \end{tabular}}
\end{center}
\end{table*}

 We assess the significance of the non-linear layer and determine the impact of its inclusion on the performance of our proposed model.
Table \ref{table:ablation} shows the performance of the proposed model on different evaluation metrics based on the non-linear layer. The performance of the proposed system improves considerably with the inclusion of non-linear layer both in the OD and demand prediction tasks as can be seen through the evaluation metrics displayed in the table.  This is because the non-linear layer captures the travelling behavior of people and provides contextual information of the place which indicates the request flow over preceding time frames. This information provides an indication of local commuter patterns in different areas which enhances the prediction accuracy of the proposed model.   % We determine the importance of the non-linear layer and evaluate the performance of the proposed system on its inclusion. %MAPE is found to exhibit greater improvement compared to MAE. % and this enhancement can be attributed to the 

\subsection{Parameter setting verification with different models and data distributions }

In this subsection, we determine the performance of the proposed model and compare it with the other models which have kept different parameters for the linear and non-linear layers. In order to check the parameter working, we have implemented these models with the GNN framework and set forth their parameters. Table \ref{table:comparison} shows the performance of the proposed model (PM) and the baselines DL \cite{Ke:ElsevierTranport_2021} FL \cite{Hu:IEEETVT_2023} with different parameter settings. As can be seen through the table the proposed model surpasses the existing models in their performance. This is because of the use of linear and non-linear dependencies to capture the patterns among requests. The proposed model determines the linear dependencies by using the data from the previous, same,  and next hours of the previous $7$ days. Concurrently, non-linear dependencies are identified through data derived from the preceding $6$ hours, which provides an accurate reflection of travel-related information while also offering insights into the contextual characteristics of the location. %The non-linear dependencies are determined by using data from the previous $6$ hours which captures the travelling data accurately and provides an indication of contextual data of the place. 
%The previous models \cite{} have only used non-linear data, or have used the data from the same hour of the previous day and the previous week to capture linear trends in data, and the data from past $2$ hours to capture non-linear trends in data. 
In contrast, previous models \cite{Ke:ElsevierTranport_2021,Hu:IEEETVT_2023}, either solely relied on non-linear data \cite{Hu:IEEETVT_2023} or exclusively considered data from the same hour of the previous day and week \cite{Ke:ElsevierTranport_2021} to capture linear trends within the data. Additionally, they \cite{Ke:ElsevierTranport_2021} limited their analysis to data from the past $2$ hours for understanding non-linear trends. These settings do not capture all the trends in underlying data, such as the travelling behavior of people or the contextual patterns of previous hours, and result in lower performance based on the evaluation metrics. The error rate is particularly high using FL \cite{Ke:ElsevierTranport_2021} since their model only utilizes the data from previous $11$ hours to predict the future demand and does not capture the linear repeating trends in data.

%Apart from verifying parameter settings on the New York dataset, we have also used the Washington DC dataset to determine if the data from the previous $6$ hours represents non-linear dependencies accurately. Table \ref{} shows the working of the proposed model with different baselines on the Washington DC dataset. It displays that the proposed model surpasses the existing baselines and uses the optimal values of parameters for linear and non-linear dependencies. It shows the working of the proposed model on different data distribution, as Washington DC has very few requests than the New York datasets. This shows that the proposed system works accurately with different distribution patterns and can be deployed in different environments.
In addition to verifying parameter settings using the New York dataset, we conducted a rigorous evaluation using the Washington DC dataset to assess the accuracy of non-linear dependencies through the data from the previous 6 hours. The findings, summarized in Table \ref{table:comparison_dc}, showcase the performance of our proposed model in comparison to various baseline models on the Washington DC dataset. These results not only demonstrate the superiority of our proposed model but also underscore its ability to optimize parameter values for both linear and non-linear dependencies. Importantly, this evaluation underscores the adaptability of our model to different data distribution patterns, particularly evident in the context of Washington DC's significantly lower request volume when contrasted with the New York dataset. %The lower count of requests in Washington DC can be attributed to the lower area covered by the Wshington DC. he lower area leads to low requests which results in smaller dataset size there in comparison to New York City which results in higher values of MAPE and MAE in Washington DC.}

%\textco{Analyzing the metrics, it becomes apparent that the error rates in the Washington DC dataset are higher than those in the New York dataset. This discrepancy primarily stems from the restricted geographic coverage of the Washington DC dataset, which encompasses a smaller area and consequently registers fewer requests. }

%Parameter verification with different data distributions
   \begin{table*}[t!]
\begin{center}
\caption{Parameter verification on New York dataset}
\label{table:comparison}
\resizebox{\textwidth}{!}{  
\begin{tabular}{ | m{6em} | m{0.6cm}| m{1.3cm} | m{1.3cm}| m{1.3cm} || m{1.3cm}| m{1.4cm} | m{1.4cm}| } 
  \hline
 Task & \makecell{Met-\\hod} & MAPE-0 & MAPE-3 & MAPE-5 & MAE-0 & MAE-3 & MAE-5 \\
  \hline
  OD & DL  FL  PM & 0.6829 0.5417 \textbf{0.3705}  & 0.9065
  0.4665 \textbf{0.3186}  & 0.9362 0.4517 \textbf{0.2962}  & 21.9279 54.9381 \textbf{4.5922}  &  60.7931 93.6856 \textbf{11.6408}  & 78.5582 106.3079 \textbf{14.6635} \\ 
  \hline
%  cell1 dummy text dummy text dummy text & cell5 & cell6 \\ 
  Demand & DL  FL  PM & 0.6829 0.4634 \textbf{0.3798}  & 0.9065 0.4472 \textbf{0.3372} &  0.9362 0.4250 \textbf{0.3183} & 21.9279 37.4200  \textbf{36.3080} &  60.7931 65.0224 \textbf{61.8143} &  78.5582 73.6000 \textbf{70.1092} \\ 
    \hline
\end{tabular}}
\end{center}
\end{table*}

   \begin{table*}[t!]
\begin{center}
\caption{Parameter verification on Washington DC dataset}
\label{table:comparison_dc}
\resizebox{\textwidth}{!}{  
\begin{tabular}{ | m{6em} | m{0.6cm}| m{1.3cm} | m{1.3cm}| m{1.3cm} || m{1.3cm}| m{1.4cm} | m{1.4cm}| } 
  \hline
 Task & \makecell{Met-\\hod} & MAPE-0 & MAPE-3 & MAPE-5 & MAE-0 & MAE-3 & MAE-5 \\
  \hline
  OD &  DL  FL PM & 0.4771 0.5685 \textbf{0.4107}  & 0.5351
  0.5745 \textbf{0.4341}  & 0.5483
  0.5581 \textbf{0.4551}  & 3.2218
  12.0491 \textbf{2.4989}  & 10.1279
  32.6828 \textbf{7.5256}  & 13.6659
  42.0089 \textbf{10.1133} \\ 
  \hline
%  cell1 dummy text dummy text dummy text & cell5 & cell6 \\ 
  Demand & DL  FL PM & 0.5523
  0.5854 \textbf{0.3798}  & 0.5713
  0.6181 \textbf{0.3681} & 0.5802
  0.6257 \textbf{0.3474} & 21.0160
  158.1532  \textbf{10.2358} &  37.9087
  272.0564 \textbf{18.0736} &  43.8728 309.5320 \textbf{20.6880} \\ 
    \hline
\end{tabular}}
\end{center}
\end{table*}

%The metrics display that the error rate is higher in the Washington DC dataset than in the New York dataset. This is primarily due to the small area covered by Washington DC datset and the lower count of requests there. In Washington DC the number of grid cells are $63$, whereas they are $361$ in New York. This displays that the area covered by Washngton DC dataset is smaller and it has fewer requests which leads to lower prediction accuracy than New York City. The lower area leads to low requests which results in smaller dataset size there in comparison to New York City which results in higher values of MAPE and MAE in Washington DC.}

\subsection{Prediction accuracy with different time slot granularities}
\begin{figure*}[t!]
    \vspace*{-43mm}
    \centering
    %\subfloat[MAE-0 for demand prediction]{%
        \includegraphics[width=0.71\textwidth]{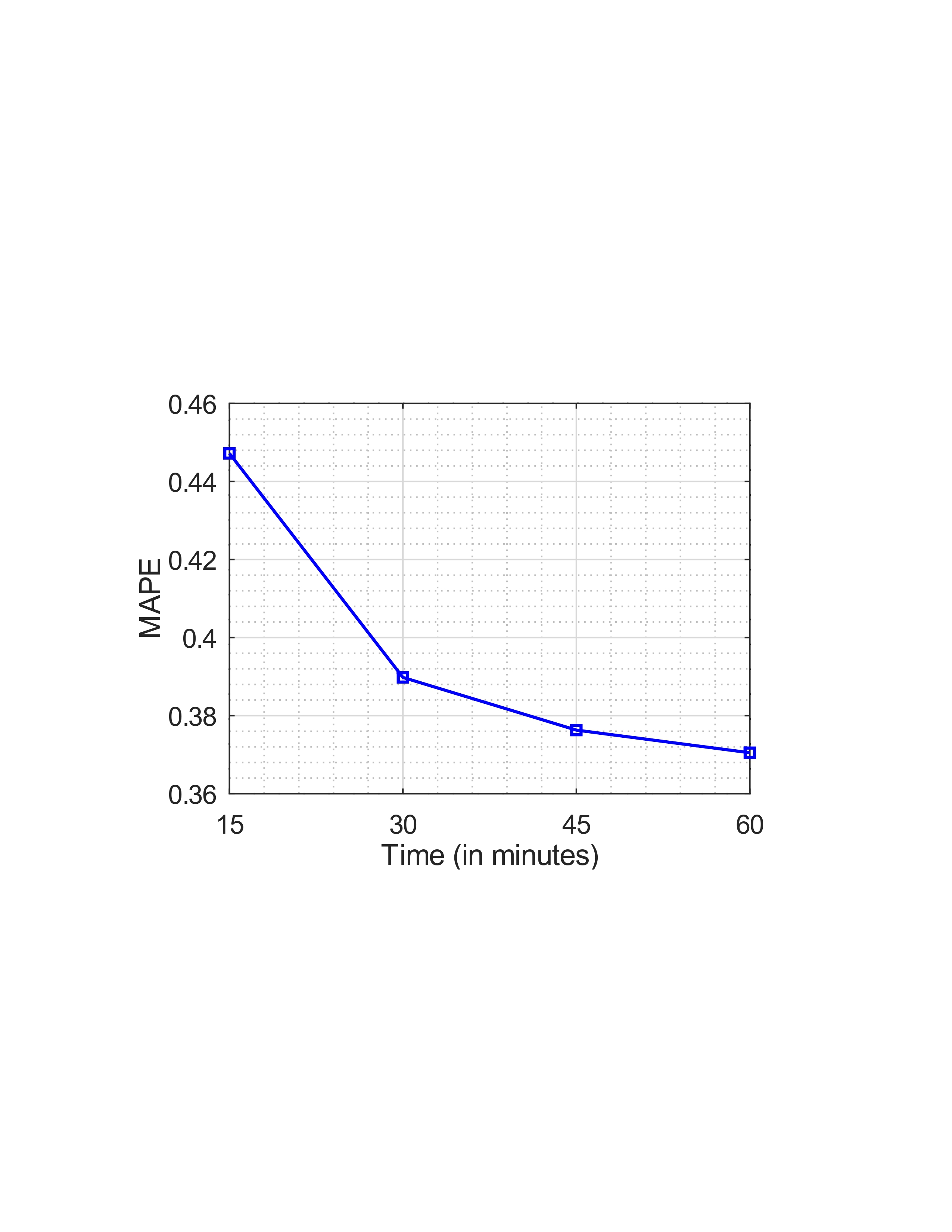}
    %}
    \vspace{-36mm}    \caption{MAPE with varying time slots}

    \label{mape_hours}
\end{figure*}

%The performance of the proposed system  depends upon the time slot granularity of prediction. 
The effectiveness of the proposed model depends upon the granularity of the time slots used for prediction. As depicted in Figure \ref{mape_hours}, the system's performance exhibits variations across different time slot granularities. Notably, the performance of the proposed model displays improvement as the time slot granularity increases. %When the model is tasked with predicting data for shorter intervals, such as $15$ minutes, it displays lower prediction accuracy.  %Figure \ref{} shows the performance with varying time slots. The performance of the proposed model improves with the increase in time slot granularity. If the model needs to predict the data for a smaller duration  like $15$ minutes the prediction accuracy will be low, and it improves with the increase in time slot granularity. 
This can be attributed to the higher data points within each time slot with the increase in time slot granularity  which results in lower prediction error. %It can be also due to lower noise in data within the higher hour data, such as due to sudden traffic congestion. With higher time slots the noise gets averaged which improves the prediction accuracy. 
Apart from this, the other reason for higher error rates in smaller time slots can be attributed to high sensitivity to short-term fluctuations and noise in the data which makes it challenging to make accurate predictions. With larger time slots this variability in data gets smoothed out, which makes it easier for the system to determine the underlying trends.

\subsection{Visual comparison of predicted and actual values}
\begin{figure*}[t!]
    \vspace*{-3mm}
    \centering
 %   \subfloat[MAE-0 for demand prediction]{%
        \includegraphics[width=0.469568\textwidth]{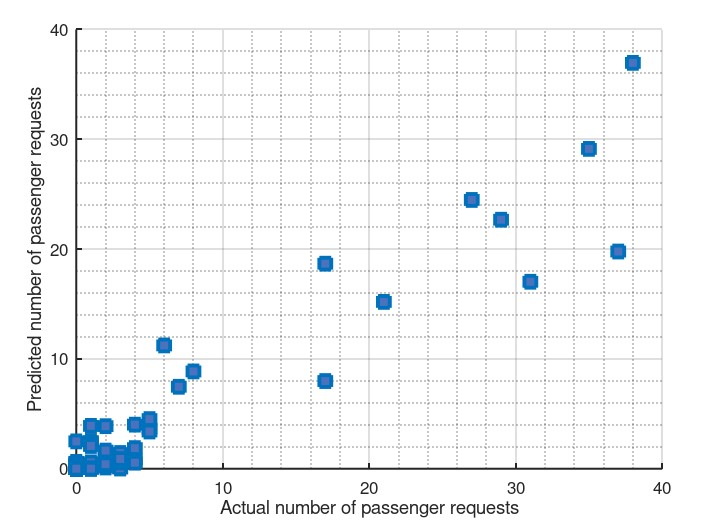}
%    }
    \vspace{-3mm}
    \caption{Predicted and actual values of passenger requests}
    \label{predicted_actual_values}
\end{figure*}

Figure \ref{predicted_actual_values} provides a visual comparison of predicted and actual values of passenger requests. %presents a correlation between predicted and actual values of passenger requests.
 In this figure, $x$-axis represents the actual number of requests, $y$ axis represents the predicted number of requests, and the plot shows the relation between the actual and predicted values. 
 As can be seen through the figure, %there is a linear correlation between the predicted and actual values and 
 the values predicted by the proposed model are closely related to the actual values. This shows that though the error rate is high and is close to $40\%$ the proposed model predicts the areas with high and low requests very well which indicates the ride-hailing platforms about the request flow in different regions. These platforms can use this data to allocate vehicles to different areas according to the demand in that area. %Moreover, this figure displays the error distribution in different areas. In the areas that are sparse in requests (which are from grid cells $1$ to $90$, and  $250$ to $361$) the predicted values are very close to actual values and are indistinguishable in the plot. This alignment is a result of the non-linear layer in the proposed model, which evaluates historical data from preceding hours to identify the low request frequency in these areas and uses it to enhance the accuracy of future predictions. %This is because the non-linear layer of the proposed model analyzes the data from the previous hours which shows that these areas were scarce in requests and uses this for the prediction of future requests. 
%For the grid cells with high demand (displayed on $x$-axis through grid cells $91$ to $249$), the proposed model predicts values with slightly less accuracy but it displays the flow of requests in different regions which can help ride-hailing platforms in directing the vehicles to these areas. The proposed model has a high error rate during these hours as the travelling patterns of passengers cannot be exactly predicted. Even though the proposed model has analyzed some aspects of behavioral patterns like the hours spent outside by people, there are various other aspects of human behavior that cannot be predicted and thereby affect the prediction accuracy of the model.  %The proposed model cannot predict exact values as }

\subsection{Comparison with previous models}
We evaluate the performance of our proposed model by comparing it with the following baselines:

\textbf{GCRN \cite{Seo:NIPS_2018}:} It captures the spatial dependencies and dynamic patterns in ride-hailing requests through GCN and RNN.

\textbf{GEML \cite{Wang:ACMKDD_2019}:} It captures the spatio-temporal dependencies in ride-hailing requests through GCN and LSTM.

\textbf{Gallat \cite{Wang:ACMTrans_2022}:} It predicts the origin and destination of requests through GNN.

\textbf{BGARN \cite{Shen_IEEEAccess:2022}:} It 
 predicts the origin and destination of requests through GNN and RNN.

   \begin{table*}[t!]
\begin{center}
\caption{Comparison on OD  and demand prediction for different methods on New York dataset}
\label{table:eval_ny}
\resizebox{\textwidth}{!}{  
\begin{tabular}{ | m{6em} | m{1.5cm}| m{1.3cm} | m{1.3cm}| m{1.3cm} || m{1.3cm}| m{1.4cm} | m{1.4cm}| } 
  \hline
 Task & \makecell{Method} & MAPE-0 & MAPE-3 & MAPE-5 & MAE-0 & MAE-3 & MAE-5 \\
  \hline
  OD &  GCRN    GEML    Gallat BGARN  PM &  0.5829 0.6732 0.6654 0.3866 \textbf{0.3705}  & 0.7853 0.8825 0.8756 0.3488 \textbf{0.3186}  & 0.8216 0.9066 0.8996 0.3269 \textbf{0.2962}  & 20.2301 19.6102 19.2130 5.4249 \textbf{4.5922}  & 57.3207  54.0873 53.9155 14.0479 \textbf{11.6408}  & 74.3299 69.7505 69.6235 17.7843 \textbf{14.6635} \\ 
  \hline
%  cell1 dummy text dummy text dummy text & cell5 & cell6 \\ 
  Demand & GCRN GEML Gallat BGARN  PM & 0.7234 0.6214 0.5872 0.4222 \textbf{0.3798}  & 0.6216 0.6203 0.5265 0.3612 \textbf{0.3372} & 0.5821  0.6277 0.4943 0.3354 \textbf{0.3183} & 105.4090 161.2817 69.3909 46.7006  \textbf{36.3080} &  184.7915 277.3443   121.4460 81.6304 \textbf{61.8143} &  209.8527  315.5562 137.7934 92.5871 \textbf{70.1092} \\ 
    \hline
\end{tabular}}
\end{center}
\end{table*}

Table \ref{table:eval_ny}  displays the performance of  GCRN, GEML, Gallat, BGARN, and our proposed model (PM) on all the evaluation parameters on the New York dataset. %As can be seen through these tables, the value of evaluation metrics is relatively higher in Washington DC. This is primarily due to the restricted geographic coverage of the Washington DC dataset which contains a small area of $63$ grid cells in comparison to $361$ grid cells covered by the New York dataset.  The smaller areas have few requests which leads to a small size of training data and results in lower prediction accuracy of the model.} %Moreover, these figures reveal interesting trends.   
The table displays that the proposed model performs better under all the metrics. This is because we have used context-aware data and analyzed the travelling pattern of users and used that to predict the future occurrence of requests. These patterns reveal the reason behind the non-linear temporal dependencies that arise in data and when these patterns are paired with the linear patterns like the morning and evening rush hours the model is found to capture different types of dependencies and thereby surpass the existing models in performance. %experimented with different values of non-linearities and determined their optimal value under various conditions. We have compared the demand as well as OD prediction, and our model is found to predict both well with an error of $0.3705$ and $0.3798$ respectively.
 % due to the dynamic nature of the service. %While certain patterns, such as the morning and evening rush hours, are predictable due to routine commuter patterns, there is no singular time frame during which users uniformly decide to travel. Consequently, this variability impacts the accuracy of the proposed model's predictions.  %Accurately predicting the time and location of passenger requests is a complex task, which involves comprehending the preferences and behaviors of passengers.%This challenge arises due to the intricate nature of humans 

\subsection{Navigating Errors: Unveiling Model Strengths Despite Challenges}

In the realm of on-demand ride-hailing services, accurately predicting the number of passenger requests at a particular time and location is a notable challenge. This challenge stems primarily from the intricate and multifaceted nature of human behavior, which is influenced by various factors. These factors encompass users' willingness to travel, local weather conditions, and traffic congestion.
While our model has successfully captured certain facets of human behavior, such as travel patterns, through its non-linear layer, there are several other behavioural patterns which are not been explored and affect the prediction accuracy of the proposed model. 
In this subsection, we will first analyze the distribution of errors and thereupon reveal the strengths of the proposed model.

\begin{figure*}[t!]
    \vspace*{-13mm}
    \centering
    %\subfloat[MAE-0 for demand prediction]{%
        \includegraphics[width=0.571\textwidth]{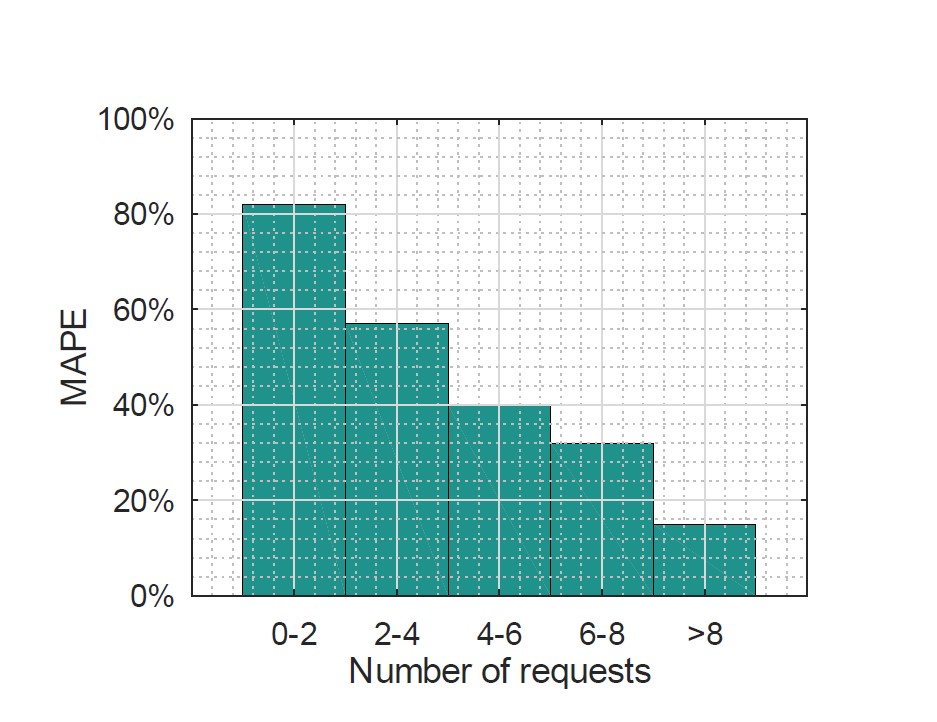}
    %}
    \vspace{-6mm}    \caption{Error distribution}

    \label{img:err_dist}
\end{figure*}

%Figure \ref{img:err_dist} shows the distribution of errors in different regions. We have categorized grid cells into different areas based upon the request count within the area. The x-axis represents the count of requests on a grid cell and y-axis represents the MAPE. It can be seen through this figure that the error decreases with the increase in number of requests that appear at a grid cell. The error rate is high when the lower request count arrives on a grid cell and this is primarily because with lower values error rate comes high and 
The histogram presented in  Figure \ref{img:err_dist} illustrates the relationship between the  MAPE and the count of requests within specific grid cells. The $x$-axis of the plot denotes the request count on a grid cell, while the $y$-axis represents the corresponding MAPE.
It can be seen through this figure that the error decreases with the increase in the number of requests that appear at a grid cell.
 When the number of requests is low for a specific grid cell, the difference between the predicted and actual values has a more pronounced impact on the calculated error rate. This is because the relative magnitude of errors is higher when the denominator (request count) is small.
In contrast, as the request count increases, the impact of individual errors diminishes relative to the total number of requests. Consequently, the error rate tends to decrease with a higher request count. This pattern is consistent with statistical principles, where larger sample sizes generally lead to more stable and representative measurements.
Therefore, the observed trend in the figure highlights the importance of considering request count in the interpretation of error rates, emphasizing that low request counts may result in disproportionately higher error rates, while larger request counts contribute to more reliable and representative error metrics.

However, it is essential to highlight that despite these inherent complexities, our proposed model has several significant strengths. %In the context of on-demand ride-hailing services, accurately predicting the number of passenger requests at a particular time and location is a notable challenge, primarily due to the intricate nature of human behavior. Several factors influence user demand, including their inclination to travel, local weather conditions at the place, and traffic congestion.Although we have captured certain aspects of human behaviour like their travelling patterns through the non-linear layer, there are several other behavioural patterns which are not been explored and affect the prediction accuracy of the proposed model. However, it is essential to highlight that despite these inherent complexities, our proposed model has several significant strengths.}

\textbf{Identification of High-Demand Areas.}

The proposed model's key strength lies in its ability to consistently identify areas with a surge in demand, even if the predicted passenger count may sometimes deviate from the actual count. This can be verified through Figure \ref{predicted_actual_values} which shows that even though the proposed model's values do not exactly match with those of actual values, it displays the demand patterns over different areas effectively. 
This capability allows the ride-hailing platforms to allocate drivers strategically to the high-demand areas, resulting in faster service, lower waiting times, and increased passenger satisfaction. % It also improves resource allocation }

\textbf{Mitigating surge pricing.}

Surge pricing emerges when demand outstrips supply. The proposed model reduces it by providing insights into areas that are likely to experience high demand. This strategic approach encourages drivers to converge around these high-demand areas, ensuring a more precise equilibrium between passenger needs and driver availability.%The proposed model, despite occasional inaccuracies, provides insights into areas that are likely to experience high demand, which concentrates the drivers around the areas where passengers are located and maintains an accurate balance of demand and supply. This results in a decrease in surge pricing %these systems can position drivers strategically to minimize wait times and enhance overall service quality.enables ride-hailing platforms to %allocate resources more effectively. By providing insights into areas likely to experience high demand, these systems can position drivers strategically to minimize wait times and enhance overall service quality.}

\textbf{Real-Time Adaptation.}

The model's flexibility in utilizing previous hour data for real-time adjustments ensures that they can respond promptly to sudden spikes in demand, even when they deviate from initial predictions. This real-time adaptability maintains service responsiveness.

\textbf{Data-Driven Decision Support.}

Beyond prediction, the proposed model serves as a valuable tool for data-driven decision-making. It offers insights into historical demand patterns, aiding in driver-passenger matching and route optimization.

In conclusion, while predicting passenger numbers in the on-demand ride-hailing industry remains a challenge, our proposed model's strengths in identifying high-demand areas, reducing surge pricing, adapting in real-time, and facilitating data-driven decisions significantly enhance the efficiency and responsiveness of the service, resulting in an improved experience for both drivers and passengers. 
Although the model may occasionally predict a higher or lower demand than the actual number of passengers arriving, it has considerable strengths in determining the areas with high demand which can facilitate in smooth allocation of resources.

\vspace{2mm}

\section{Applications}
\label{sec:apps}
%Ride-hailing services generate an enormous volume of passenger data everyday. This data can be used effectively for predicting the future occurences of events. In this paper, we have used the GNN based model to predict the future demand that can arise at different locations. Moreover, we have predicted the future origin and destination of ride-hailing requests

%The advent of ride-hailing services has resulted in a massive volume of daily passenger data, presenting an excellent opportunity for predictive analysis. 
Ride-hailing services generate an enormous volume of passenger data every day. This data can be used effectively for predicting the future occurrences of passenger requests. 
This paper focuses on harnessing the power of GNN  to make  predictions regarding future origin and destinations of requests.  The applications of such predictions are diverse and impactful. Some of them include:

\textbf{Revenue optimization and passenger satisfaction:} When the ride-hailing platforms are provided with data about the passenger's origin and destination in advance, they can use this data to strategically dispatch drivers to the areas that will have high demand which results in efficient coverage of passenger requests. %This results in higher revenue for these platforms due to the higher passenger orders taken by the platform.  
This approach leads to an increase in revenue of the platform as they are able to accommodate a larger number of passenger orders due to the presence of drivers in passenger-centric areas.
%Moreover, it  also results in passenger satisfaction as their waiting time gets decreased due to the dispatching of drivers to high demand areas. The passengers will get the ride quickly due to the higher supply in the customer-centric areas.

This data-driven approach also has a significant impact on passenger satisfaction. By dispatching drivers to areas with high demand, waiting times for passengers are reduced. As a result, passengers experience shorter wait times and are able to secure rides promptly, benefiting from the higher supply of drivers in high-demand regions. %This alignment of driver deployment with passenger demand not only enhances operational efficiency but also contributes to overall passenger satisfaction.

\textbf{Fleet Management:} 
Fleet size refers to the number of vehicles required to service passenger requests that can arrive in a given area. By predicting the future origins and destinations of requests, ride-hailing companies get an advanced estimate of the vehicles required to cover the requests that arrive over different frames of the day. 
This will result in efficient resource utilization as the companies will use the higher resources when the demand is expected to be high and will constrain the resources over the time periods when the demand is expected to be low.
%By getting the estimate of passenger demand over different times of the day, ride-hailing companies can determine the number of vehicles required to service all of their requests. 

\textbf{Eco-friendly rides:}
Ridesharing services result in eco-friendly rides as they pair multiple passengers with similar routes in a single vehicle. This results in efficient utilization of vehicles and decreases the count of vehicles on the road. However, the current statistics display that only $15\%$ of rides are shared \cite{Article:UberPollution_2020}, whereas $73\%$ of rides can be shared with a slight increase in the waiting time of passengers \cite{Cai:ElsevierEnergy_2019}. This highlights the substantial scope for improving ridesharing rates if accurate predictions of passenger origins and destinations are made in advance. By accurately forecasting the pickup and drop-off points of passenger requests, ridesharing platforms can proactively match passengers with similar routes, thereby increasing the likelihood of shared rides. %The higher percentage of rides can be shared  if the origin as well as destination of passenger requests is predicted beforehand.
%the passengers are paired effectively in a single vehicle, which can be done 

\textbf{Traffic Planning:} The predicted future demand and ride origins/destinations can contribute to traffic planning efforts. By analyzing these patterns, transportation authorities can gain insights into congestion-prone areas and make informed decisions regarding infrastructure improvements, traffic flow management, and public transportation planning.

%\textbf{Surge Pricing:} Accurate predictions of future demand allow ride-hailing platforms to implement surge pricing mechanisms effectively. By identifying areas likely to experience high demand, surge pricing can be applied proactively to incentivize drivers to be available in those locations, maintaining a balance between supply and demand.

%\textbf{Service Expansion:} Predictive models can assist ride-hailing companies in identifying areas with emerging demand. By forecasting future ride-hailing needs at different locations, providers can strategically expand their services into new markets or adjust existing service coverage to cater to growing demand.

Overall, the application of GNN-based models for predicting future demand, origins, and destinations offers numerous opportunities for improving operational efficiency, enhancing customer experience, and optimizing resource allocation in the dynamic landscape of ride-hailing services.

\section{Conclusion}
\label{sec:conclusion}
In this paper, we define passenger mobility through origin-destination prediction which is helpful in directing optimal routes to drivers and providing matching algorithms for drivers and passengers. %, which enhances the functionality of service providers.  
OD prediction is complex in comparison to demand prediction as it predicts the demand in a certain region and the destination of these demands. We have used GNN to capture the spatio-temporal dependencies among requests and predicted the OD pair of requests. %We have particularly focused on temporal non-linearities that cause the dependencies among requests and can be used to follow the patterns apart from the regular repeating patterns in data.
We have particularly focused on temporal non-linearities that arise due to contextual events or the travelling patterns of people.  While modelling these dependencies, the length of the grid cell is an important parameter and it can regulate the neighborhood count. We have determined its value through extensive simulations. The parameters that decide the grid cell length and the non-linearities are estimated for demand prediction models also. Extensive simulations determine superior performance by our proposed model as compared to the existing baselines.

%\section{Declarations}

%\section*{Disclosure statement}

%The authors report there are no competing interests to declare

%\section*{Funding}

%No support was provided for this research. 

%\section*{Availability of data and materials }
%The data can be made available on request.

%\section*{Notes on contributor(s)}

%Aqsa Ashraf Makhdomi was responsible for conducting the experiments, collecting and analyzing the data, and writing the initial draft of the manuscript.

%Iqra Altaf Gillani contributed to the study design, reviewed the manuscript, and provided critical feedback and expertise in interpreting the results.

\bibliographystyle{apacite}
\bibliography{interactapasample}

\begin{thebibliography}{}

\bibitem [\protect \citeauthoryear {%
Cai%
, Wang%
, Adriaens%
\BCBL {}\ \BBA {} Xu%
}{%
Cai%
\ \protect \BOthers {.}}{%
{\protect \APACyear {2019}}%
}]{%
Cai:ElsevierEnergy_2019}
\APACinsertmetastar {%
Cai:ElsevierEnergy_2019}%
\begin{APACrefauthors}%
Cai, H.%
, Wang, X.%
, Adriaens, P.%
\BCBL {}\ \BBA {} Xu, M.%
\end{APACrefauthors}%
\unskip\
\newblock
\APACrefYearMonthDay{2019}{}{}.
\newblock
{\BBOQ}\APACrefatitle {Environmental benefits of taxi ride sharing in Beijing}
  {Environmental benefits of taxi ride sharing in beijing}.{\BBCQ}
\newblock
\APACjournalVolNumPages{Energy}{174}{}{503-508}.
\PrintBackRefs{\CurrentBib}

\bibitem [\protect \citeauthoryear {%
Cho%
, Myers%
\BCBL {}\ \BBA {} Leskovec%
}{%
Cho%
\ \protect \BOthers {.}}{%
{\protect \APACyear {2011}}%
}]{%
Cho:KDD_2011}
\APACinsertmetastar {%
Cho:KDD_2011}%
\begin{APACrefauthors}%
Cho, E.%
, Myers, S\BPBI A.%
\BCBL {}\ \BBA {} Leskovec, J.%
\end{APACrefauthors}%
\unskip\
\newblock
\APACrefYearMonthDay{2011}{}{}.
\newblock
{\BBOQ}\APACrefatitle {Friendship and mobility: user movement in location-based
  social networks} {Friendship and mobility: user movement in location-based
  social networks}.{\BBCQ}
\newblock
\BIn{} \APACrefbtitle {Proceedings of the 17th ACM SIGKDD international
  conference on Knowledge discovery and data mining} {Proceedings of the 17th
  acm sigkdd international conference on knowledge discovery and data mining}\
  (\BPGS\ 1082--1090).
\PrintBackRefs{\CurrentBib}

\bibitem [\protect \citeauthoryear {%
de Souza%
, Pedrosa%
, Botega%
\BCBL {}\ \BBA {} Villas%
}{%
de Souza%
\ \protect \BOthers {.}}{%
{\protect \APACyear {2018}}%
}]{%
desouza:IEEEConf_2018}
\APACinsertmetastar {%
desouza:IEEEConf_2018}%
\begin{APACrefauthors}%
de Souza, A\BPBI M.%
, Pedrosa, L\BPBI L\BPBI C.%
, Botega, L\BPBI C.%
\BCBL {}\ \BBA {} Villas, L.%
\end{APACrefauthors}%
\unskip\
\newblock
\APACrefYearMonthDay{2018}{June}{}.
\newblock
{\BBOQ}\APACrefatitle {Itssafe: An Intelligent Transportation System for
  Improving Safety and Traffic Efficiency} {Itssafe: An intelligent
  transportation system for improving safety and traffic efficiency}.{\BBCQ}
\newblock
\BIn{} \APACrefbtitle {2018 IEEE 87th Vehicular Technology Conference (VTC
  Spring)} {2018 ieee 87th vehicular technology conference (vtc spring)}\
  (\BPG~1-7).
\PrintBackRefs{\CurrentBib}

\bibitem [\protect \citeauthoryear {%
Diffey%
}{%
Diffey%
}{%
{\protect \APACyear {2011}}%
}]{%
diffey_britishjournal:2011}
\APACinsertmetastar {%
diffey_britishjournal:2011}%
\begin{APACrefauthors}%
Diffey, B\BPBI L.%
\end{APACrefauthors}%
\unskip\
\newblock
\APACrefYearMonthDay{2011}{}{}.
\newblock
{\BBOQ}\APACrefatitle {An overview analysis of the time people spend outdoors}
  {An overview analysis of the time people spend outdoors}.{\BBCQ}
\newblock
\APACjournalVolNumPages{British Journal of Dermatology}{164}{4}{848--854}.
\PrintBackRefs{\CurrentBib}

\bibitem [\protect \citeauthoryear {%
Galbrun%
, Pelechrinis%
\BCBL {}\ \BBA {} Terzi%
}{%
Galbrun%
\ \protect \BOthers {.}}{%
{\protect \APACyear {2016}}%
}]{%
Galbrun:ElsevierInforSys_2016}
\APACinsertmetastar {%
Galbrun:ElsevierInforSys_2016}%
\begin{APACrefauthors}%
Galbrun, E.%
, Pelechrinis, K.%
\BCBL {}\ \BBA {} Terzi, E.%
\end{APACrefauthors}%
\unskip\
\newblock
\APACrefYearMonthDay{2016}{}{}.
\newblock
{\BBOQ}\APACrefatitle {Urban navigation beyond shortest route: The case of safe
  paths} {Urban navigation beyond shortest route: The case of safe
  paths}.{\BBCQ}
\newblock
\APACjournalVolNumPages{Information Systems}{57}{}{160-171}.
\PrintBackRefs{\CurrentBib}

\bibitem [\protect \citeauthoryear {%
Gao%
, Li%
, Wang%
\BCBL {}\ \BBA {} Huang%
}{%
Gao%
\ \protect \BOthers {.}}{%
{\protect \APACyear {2022}}%
}]{%
Gao:IEEETrans_2022}
\APACinsertmetastar {%
Gao:IEEETrans_2022}%
\begin{APACrefauthors}%
Gao, J.%
, Li, X.%
, Wang, C.%
\BCBL {}\ \BBA {} Huang, X.%
\end{APACrefauthors}%
\unskip\
\newblock
\APACrefYearMonthDay{2022}{Aug}{}.
\newblock
{\BBOQ}\APACrefatitle {BM-DDPG: An Integrated Dispatching Framework for
  Ride-Hailing Systems} {Bm-ddpg: An integrated dispatching framework for
  ride-hailing systems}.{\BBCQ}
\newblock
\APACjournalVolNumPages{IEEE Transactions on Intelligent Transportation
  Systems}{23}{8}{11666-11676}.
\PrintBackRefs{\CurrentBib}

\bibitem [\protect \citeauthoryear {%
Garg%
\ \BBA {} Ranu%
}{%
Garg%
\ \BBA {} Ranu%
}{%
{\protect \APACyear {2018}}%
}]{%
Garg:ACMKDD_2018}
\APACinsertmetastar {%
Garg:ACMKDD_2018}%
\begin{APACrefauthors}%
Garg, N.%
\BCBT {}\ \BBA {} Ranu, S.%
\end{APACrefauthors}%
\unskip\
\newblock
\APACrefYearMonthDay{2018}{}{}.
\newblock
{\BBOQ}\APACrefatitle {Route Recommendations for Idle Taxi Drivers: Find Me the
  Shortest Route to a Customer!} {Route recommendations for idle taxi drivers:
  Find me the shortest route to a customer!}{\BBCQ}
\newblock
\BIn{} \APACrefbtitle {Proceedings of the 24th ACM SIGKDD International
  Conference on Knowledge Discovery \& Data Mining} {Proceedings of the 24th
  acm sigkdd international conference on knowledge discovery \& data mining}\
  (\BPG~1425–1434).
\newblock
\APACaddressPublisher{New York, NY, USA}{Association for Computing Machinery}.
\PrintBackRefs{\CurrentBib}

\bibitem [\protect \citeauthoryear {%
Hamilton%
, Ying%
\BCBL {}\ \BBA {} Leskovec%
}{%
Hamilton%
\ \protect \BOthers {.}}{%
{\protect \APACyear {2017}}%
}]{%
Hamilton:neuips_2017}
\APACinsertmetastar {%
Hamilton:neuips_2017}%
\begin{APACrefauthors}%
Hamilton, W.%
, Ying, Z.%
\BCBL {}\ \BBA {} Leskovec, J.%
\end{APACrefauthors}%
\unskip\
\newblock
\APACrefYearMonthDay{2017}{}{}.
\newblock
{\BBOQ}\APACrefatitle {Inductive representation learning on large graphs}
  {Inductive representation learning on large graphs}.{\BBCQ}
\newblock
\APACjournalVolNumPages{Advances in neural information processing
  systems}{30}{}{}.
\PrintBackRefs{\CurrentBib}

\bibitem [\protect \citeauthoryear {%
Han%
, Song%
, Wang%
\BCBL {}\ \protect \BOthers {.}}{%
Han%
\ \protect \BOthers {.}}{%
{\protect \APACyear {2004}}%
}]{%
Han_2004}
\APACinsertmetastar {%
Han_2004}%
\begin{APACrefauthors}%
Han, C.%
, Song, S.%
, Wang, C.%
\BCBL {}\ \BOthersPeriod {.}\end{APACrefauthors}%
\unskip\
\newblock
\APACrefYearMonthDay{2004}{}{}.
\newblock
{\BBOQ}\APACrefatitle {A real-time short-term traffic flow adaptive forecasting
  method based on ARIMA model} {A real-time short-term traffic flow adaptive
  forecasting method based on arima model}.{\BBCQ}
\newblock
\APACjournalVolNumPages{Journal of system simulation}{16}{7}{1530--1535}.
\PrintBackRefs{\CurrentBib}

\bibitem [\protect \citeauthoryear {%
Hu%
\ \protect \BOthers {.}}{%
Hu%
\ \protect \BOthers {.}}{%
{\protect \APACyear {2023}}%
}]{%
Hu:IEEETVT_2023}
\APACinsertmetastar {%
Hu:IEEETVT_2023}%
\begin{APACrefauthors}%
Hu, S.%
, Ye, Y.%
, Hu, Q.%
, Liu, X.%
, Cao, S.%
, Yang, H\BPBI H.%
\BDBL {}Wu, C.%
\end{APACrefauthors}%
\unskip\
\newblock
\APACrefYearMonthDay{2023}{}{}.
\newblock
{\BBOQ}\APACrefatitle {A Federated Learning-Based Framework for Ride-sourcing
  Traffic Demand Prediction} {A federated learning-based framework for
  ride-sourcing traffic demand prediction}.{\BBCQ}
\newblock
\APACjournalVolNumPages{IEEE Transactions on Vehicular Technology}{}{}{1-15}.
\newblock
\begin{APACrefDOI} \doi{10.1109/TVT.2023.3287221} \end{APACrefDOI}
\PrintBackRefs{\CurrentBib}

\bibitem [\protect \citeauthoryear {%
Jin%
\ \protect \BOthers {.}}{%
Jin%
\ \protect \BOthers {.}}{%
{\protect \APACyear {2020}}%
}]{%
Jin:ElsevierTransp_2020}
\APACinsertmetastar {%
Jin:ElsevierTransp_2020}%
\begin{APACrefauthors}%
Jin, G.%
, Cui, Y.%
, Zeng, L.%
, Tang, H.%
, Feng, Y.%
\BCBL {}\ \BBA {} Huang, J.%
\end{APACrefauthors}%
\unskip\
\newblock
\APACrefYearMonthDay{2020}{}{}.
\newblock
{\BBOQ}\APACrefatitle {Urban ride-hailing demand prediction with multiple
  spatio-temporal information fusion network} {Urban ride-hailing demand
  prediction with multiple spatio-temporal information fusion network}.{\BBCQ}
\newblock
\APACjournalVolNumPages{Transportation Research Part C: Emerging
  Technologies}{117}{}{102665}.
\PrintBackRefs{\CurrentBib}

\bibitem [\protect \citeauthoryear {%
Ke%
\ \protect \BOthers {.}}{%
Ke%
\ \protect \BOthers {.}}{%
{\protect \APACyear {2021}}%
}]{%
Ke:ElsevierTranport_2021}
\APACinsertmetastar {%
Ke:ElsevierTranport_2021}%
\begin{APACrefauthors}%
Ke, J.%
, Qin, X.%
, Yang, H.%
, Zheng, Z.%
, Zhu, Z.%
\BCBL {}\ \BBA {} Ye, J.%
\end{APACrefauthors}%
\unskip\
\newblock
\APACrefYearMonthDay{2021}{}{}.
\newblock
{\BBOQ}\APACrefatitle {Predicting origin-destination ride-sourcing demand with
  a spatio-temporal encoder-decoder residual multi-graph convolutional network}
  {Predicting origin-destination ride-sourcing demand with a spatio-temporal
  encoder-decoder residual multi-graph convolutional network}.{\BBCQ}
\newblock
\APACjournalVolNumPages{Transportation Research Part C: Emerging
  Technologies}{122}{}{102858}.
\PrintBackRefs{\CurrentBib}

\bibitem [\protect \citeauthoryear {%
Lippi%
, Bertini%
\BCBL {}\ \BBA {} Frasconi%
}{%
Lippi%
\ \protect \BOthers {.}}{%
{\protect \APACyear {2013}}%
}]{%
Lippi:IEEETrans-2013}
\APACinsertmetastar {%
Lippi:IEEETrans-2013}%
\begin{APACrefauthors}%
Lippi, M.%
, Bertini, M.%
\BCBL {}\ \BBA {} Frasconi, P.%
\end{APACrefauthors}%
\unskip\
\newblock
\APACrefYearMonthDay{2013}{June}{}.
\newblock
{\BBOQ}\APACrefatitle {Short-Term Traffic Flow Forecasting: An Experimental
  Comparison of Time-Series Analysis and Supervised Learning} {Short-term
  traffic flow forecasting: An experimental comparison of time-series analysis
  and supervised learning}.{\BBCQ}
\newblock
\APACjournalVolNumPages{IEEE Transactions on Intelligent Transportation
  Systems}{14}{2}{871-882}.
\PrintBackRefs{\CurrentBib}

\bibitem [\protect \citeauthoryear {%
Ortiz-Ospina%
}{%
Ortiz-Ospina%
}{%
{\protect \APACyear {2020}}%
}]{%
Article:workhours}
\APACinsertmetastar {%
Article:workhours}%
\begin{APACrefauthors}%
Ortiz-Ospina, E.%
\end{APACrefauthors}%
\unskip\
\newblock
\APACrefYearMonthDay{2020}{}{}.
\newblock
\APACrefbtitle {How do people across the world spend their time and what does
  this tell us about living conditions?} {How do people across the world spend
  their time and what does this tell us about living conditions?}
\newblock
\APAChowpublished
  {\url{https://ourworldindata.org/time-use-living-conditions}}.
\newblock
\APACrefnote{Accessed: 2022-10-13}
\PrintBackRefs{\CurrentBib}

\bibitem [\protect \citeauthoryear {%
Petit%
}{%
Petit%
}{%
{\protect \APACyear {2020}}%
}]{%
Article:UberPollution_2020}
\APACinsertmetastar {%
Article:UberPollution_2020}%
\begin{APACrefauthors}%
Petit, Y\BPBI L.%
\end{APACrefauthors}%
\unskip\
\newblock
\APACrefYearMonthDay{2020}{}{}.
\newblock
\APACrefbtitle {Uber pollutes more than the cars it replaces–US scientists.}
  {Uber pollutes more than the cars it replaces–us scientists.}
\newblock
\APAChowpublished
  {\url{https://www.transportenvironment.org/discover/uber-pollutes-more-cars-it-replaces-us-scientists/}}.
\newblock
\APACrefnote{Accessed: 2022-02-28}
\PrintBackRefs{\CurrentBib}

\bibitem [\protect \citeauthoryear {%
Schaller%
}{%
Schaller%
}{%
{\protect \APACyear {2021}}%
}]{%
Schaller:ElsevierTransport_2021}
\APACinsertmetastar {%
Schaller:ElsevierTransport_2021}%
\begin{APACrefauthors}%
Schaller, B.%
\end{APACrefauthors}%
\unskip\
\newblock
\APACrefYearMonthDay{2021}{}{}.
\newblock
{\BBOQ}\APACrefatitle {Can sharing a ride make for less traffic? Evidence from
  Uber and Lyft and implications for cities} {Can sharing a ride make for less
  traffic? evidence from uber and lyft and implications for cities}.{\BBCQ}
\newblock
\APACjournalVolNumPages{Transport Policy}{102}{}{1-10}.
\PrintBackRefs{\CurrentBib}

\bibitem [\protect \citeauthoryear {%
Seo%
, Defferrard%
, Vandergheynst%
\BCBL {}\ \BBA {} Bresson%
}{%
Seo%
\ \protect \BOthers {.}}{%
{\protect \APACyear {2018}}%
}]{%
Seo:NIPS_2018}
\APACinsertmetastar {%
Seo:NIPS_2018}%
\begin{APACrefauthors}%
Seo, Y.%
, Defferrard, M.%
, Vandergheynst, P.%
\BCBL {}\ \BBA {} Bresson, X.%
\end{APACrefauthors}%
\unskip\
\newblock
\APACrefYearMonthDay{2018}{}{}.
\newblock
{\BBOQ}\APACrefatitle {Structured sequence modeling with graph convolutional
  recurrent networks} {Structured sequence modeling with graph convolutional
  recurrent networks}.{\BBCQ}
\newblock
\BIn{} \APACrefbtitle {Neural Information Processing: 25th International
  Conference, ICONIP 2018, Siem Reap, Cambodia, December 13-16, 2018,
  Proceedings, Part I 25} {Neural information processing: 25th international
  conference, iconip 2018, siem reap, cambodia, december 13-16, 2018,
  proceedings, part i 25}\ (\BPGS\ 362--373).
\PrintBackRefs{\CurrentBib}

\bibitem [\protect \citeauthoryear {%
Shan%
, Zhou%
\BCBL {}\ \BBA {} Wang%
}{%
Shan%
\ \protect \BOthers {.}}{%
{\protect \APACyear {2018}}%
}]{%
Shan:IEEEConf_2018}
\APACinsertmetastar {%
Shan:IEEEConf_2018}%
\begin{APACrefauthors}%
Shan, D.%
, Zhou, W.%
\BCBL {}\ \BBA {} Wang, J.%
\end{APACrefauthors}%
\unskip\
\newblock
\APACrefYearMonthDay{2018}{July}{}.
\newblock
{\BBOQ}\APACrefatitle {A Novel Personalized Dynamic Route Recommendation
  Approach Based on Pearson Similarity Coefficient in Cooperative
  Vehicle-Infrastructure Systems} {A novel personalized dynamic route
  recommendation approach based on pearson similarity coefficient in
  cooperative vehicle-infrastructure systems}.{\BBCQ}
\newblock
\BIn{} \APACrefbtitle {2018 IEEE 8th Annual International Conference on CYBER
  Technology in Automation, Control, and Intelligent Systems (CYBER)} {2018
  ieee 8th annual international conference on cyber technology in automation,
  control, and intelligent systems (cyber)}\ (\BPG~1270-1275).
\PrintBackRefs{\CurrentBib}

\bibitem [\protect \citeauthoryear {%
Shen%
, Tziritas%
\BCBL {}\ \BBA {} Theodoropoulos%
}{%
Shen%
\ \protect \BOthers {.}}{%
{\protect \APACyear {2022}}%
}]{%
Shen_IEEEAccess:2022}
\APACinsertmetastar {%
Shen_IEEEAccess:2022}%
\begin{APACrefauthors}%
Shen, J.%
, Tziritas, N.%
\BCBL {}\ \BBA {} Theodoropoulos, G.%
\end{APACrefauthors}%
\unskip\
\newblock
\APACrefYearMonthDay{2022}{}{}.
\newblock
{\BBOQ}\APACrefatitle {A Baselined Gated Attention Recurrent Network for
  Request Prediction in Ridesharing} {A baselined gated attention recurrent
  network for request prediction in ridesharing}.{\BBCQ}
\newblock
\APACjournalVolNumPages{IEEE Access}{10}{}{86423--86434}.
\PrintBackRefs{\CurrentBib}

\bibitem [\protect \citeauthoryear {%
Wang%
\ \protect \BOthers {.}}{%
Wang%
\ \protect \BOthers {.}}{%
{\protect \APACyear {2019}}%
}]{%
Wang:ACMKDD_2019}
\APACinsertmetastar {%
Wang:ACMKDD_2019}%
\begin{APACrefauthors}%
Wang, Y.%
, Yin, H.%
, Chen, H.%
, Wo, T.%
, Xu, J.%
\BCBL {}\ \BBA {} Zheng, K.%
\end{APACrefauthors}%
\unskip\
\newblock
\APACrefYearMonthDay{2019}{}{}.
\newblock
{\BBOQ}\APACrefatitle {Origin-Destination Matrix Prediction via Graph
  Convolution: A New Perspective of Passenger Demand Modeling}
  {Origin-destination matrix prediction via graph convolution: A new
  perspective of passenger demand modeling}.{\BBCQ}
\newblock
\BIn{} \APACrefbtitle {Proceedings of the 25th ACM SIGKDD International
  Conference on Knowledge Discovery \& Data Mining} {Proceedings of the 25th
  acm sigkdd international conference on knowledge discovery \& data mining}\
  (\BPG~1227–1235).
\newblock
\APACaddressPublisher{New York, NY, USA}{Association for Computing Machinery}.
\PrintBackRefs{\CurrentBib}

\bibitem [\protect \citeauthoryear {%
Wang%
\ \protect \BOthers {.}}{%
Wang%
\ \protect \BOthers {.}}{%
{\protect \APACyear {2021}}%
{\protect \APACexlab {{\protect \BCnt {1}}}}}]{%
Wang:IEEEConf_2021}
\APACinsertmetastar {%
Wang:IEEEConf_2021}%
\begin{APACrefauthors}%
Wang, Y.%
, Yin, H.%
, Chen, T.%
, Liu, C.%
, Wang, B.%
, Wo, T.%
\BCBL {}\ \BBA {} Xu, J.%
\end{APACrefauthors}%
\unskip\
\newblock
\APACrefYearMonthDay{2021{\protect \BCnt {1}}}{}{}.
\newblock
{\BBOQ}\APACrefatitle {Gallat: A Spatiotemporal Graph Attention Network for
  Passenger Demand Prediction} {Gallat: A spatiotemporal graph attention
  network for passenger demand prediction}.{\BBCQ}
\newblock
\BIn{} \APACrefbtitle {2021 IEEE 37th International Conference on Data
  Engineering (ICDE)} {2021 ieee 37th international conference on data
  engineering (icde)}\ (\BPG~2129-2134).
\PrintBackRefs{\CurrentBib}

\bibitem [\protect \citeauthoryear {%
Wang%
\ \protect \BOthers {.}}{%
Wang%
\ \protect \BOthers {.}}{%
{\protect \APACyear {2021}}%
{\protect \APACexlab {{\protect \BCnt {2}}}}}]{%
Wang:ACMTrans_2022}
\APACinsertmetastar {%
Wang:ACMTrans_2022}%
\begin{APACrefauthors}%
Wang, Y.%
, Yin, H.%
, Chen, T.%
, Liu, C.%
, Wang, B.%
, Wo, T.%
\BCBL {}\ \BBA {} Xu, J.%
\end{APACrefauthors}%
\unskip\
\newblock
\APACrefYearMonthDay{2021{\protect \BCnt {2}}}{nov}{}.
\newblock
{\BBOQ}\APACrefatitle {Passenger Mobility Prediction via Representation
  Learning for Dynamic Directed and Weighted Graphs} {Passenger mobility
  prediction via representation learning for dynamic directed and weighted
  graphs}.{\BBCQ}
\newblock
\APACjournalVolNumPages{ACM Trans. Intell. Syst. Technol.}{13}{1}{}.
\PrintBackRefs{\CurrentBib}

\end{thebibliography}

\end{document}